\documentclass[journal,twoside,10pt]{IEEEtran}

\usepackage[T1]{fontenc}
\usepackage[utf8]{inputenc}
\usepackage{amsmath,amssymb,amsfonts}
\usepackage{algorithmic}
\usepackage{graphicx}
\usepackage{booktabs}
\usepackage{multirow}
\usepackage{array}
\usepackage{url}
\usepackage{hyperref}
\usepackage{cite}
\usepackage{balance}
\usepackage{tabularx}
\usepackage{subcaption}
\usepackage{makecell}
\usepackage[font=small,skip=2pt]{caption}

\hypersetup{
  colorlinks = true,
  linkcolor  = black,
  citecolor  = black,
  urlcolor   = black
}

\newcolumntype{C}[1]{>{\centering\arraybackslash}p{#1}}

\title{Dual-Temporal LSTM with Hybrid Attention for Airline Passenger Load Factor Forecasting: Integrating Intra-Flight and Inter-Flight Booking Dynamics}

\author{A~S~M~Nazrul~Islam,~\IEEEmembership{}
        Md.~Hasanul~Kabir,~\IEEEmembership{}
        ~Md.~Liakot~Ali,~\IEEEmembership{}
        and~Joydeb~Kumar~Sana%
\thanks{Manuscript received [DATE]; revised [DATE]; accepted [DATE].
        Date of publication [DATE]; date of current version [DATE].
        \textit{Corresponding author: Md. Liakot Ali, and Joydeb Kumar Sana}}%
\thanks{Md.Liakot Ali (e-mail: liakot@iict.buet.ac.bd), A.S.M.Nazrul Islam (e-mail:asmnazrul1363@gmail.com) and Joydeb Kumar Sana (e-mail: joydebsana@iict.buet.ac.bd, joysana@gmail.com)  are with the Institute of Information and Communication Technology,
        Bangladesh University of Engineering and Technology,
        Dhaka, Bangladesh.
        Md. Hasanul Kabir (email: hasanul@iut-dhaka.edu) is with the Islamic University of
        Technology, Dhaka, Bangladesh.
         }%
\thanks{ }%
}

\begin{document}
\maketitle

% ============================================================
%  ABSTRACT
% ============================================================
\begin{abstract}
Accurate short-term demand forecasting is crucial to airline revenue management, yet most existing systems fail to meet this need because current models treat booking data as a single temporal dimension, either the accumulation of bookings for a specific flight or the historical booking profile of the same route. This unidimensional view discards information carried by the other temporal stream and forecasting absolute passenger counts introduces a further operational fragility when change in planned aircraft type alters total seat capacity. This study addresses both limitations. A dual-stream Long Short-Term Memory (LSTM) integrated with attention framework is proposed that simultaneously processes two complementary input sequences: a horizontal sequence capturing intra-flight booking accumulation over the days preceding departure, and a vertical sequence capturing inter-flight booking patterns at fixed days-before-departure offsets across historical flights. A novel offset-based algorithm constructs the vertical sequence, and a seven-stage feature selection pipeline reduces 39 engineered features to 17 optimally informative inputs. Rather than predicting absolute passenger counts, the model targets Passenger Load Factor (PLF); the ratio of booked seats to total capacity, which remains stable across aircraft type changes. Multiple dual-stream architectural variants, combining self-attention, cross-attention, and hybrid attention with concatenation, residual, and gated fusion strategies, are developed and evaluated. Experiments on real-world reservation data from the national airline of Bangladesh, Biman Bangladesh Airlines (BBA), demonstrate that the proposed hybrid model achieves a Mean Absolute Error of 2.8167 and a coefficient of determination ($R^{2}$) of 0.9495, outperforming single-stream baselines, tree-based models, and three prior dual-LSTM architectures applied to the same data. Validation across four flight category pairs; domestic versus international, direct versus transit, high versus low frequency, and short versus mid versus long haul confirms that the model generalizes across operationally diverse route types. Biman Bangladesh Airlines (BBA) has officially integrated this methodology into its operations.
\end{abstract}

% ============================================================
%  INDEX TERMS
% ============================================================

% ============================================================
%  I. INTRODUCTION
% ============================================================
\section{Introduction}

\subsection{Background}

The airline industry operates on margins that leave little room for
forecasting error. With net profit per passenger hovering near \$2.60
industry-wide in 2025~\cite{iata2025}, and passenger revenue
accounting for approximately 72\% of total airline income, the
accuracy of demand forecasts directly determines the quality of every
downstream revenue management decision, from seat inventory allocation
and overbooking limits to dynamic pricing and fleet assignment. A
forecast that overestimates demand leads to aggressive pricing that
drives away late-booking passengers; whereas one that underestimates
it leaves revenue on the table through underpriced inventory. In
either case, the cost of opportunity is immediate.

In flight level demand forecasting, the main concern is how many seats
on a specific flight, on a specific date, on a specific route, will
ultimately be occupied. The data that addresses this question is the
booking curve; the accumulation of confirmed reservations from the
time a flight opens for sale, up to the departure date itself. The
curve does not exhibit simple monotonic growth. Rather it reflects
competitive pricing changes, promotional events, holiday effects,
last-minute business travel, and route-specific booking behaviors that
vary systematically across flight types and seasons.

Traditional approaches to modelling this curve have relied on
statistical methods like, autoregressive integrated moving average
(ARIMA) models~\cite{nieto2018}, and regression-based
frameworks~\cite{suryan2017}. These methods provided solutions under
stationarity assumptions but struggle when booking dynamics are
non-linear or exhibit irregular seasonality. The gradual displacement
of these methods by machine learning and deep learning approaches
reflects a broader recognition that the temporal structure of airline
booking data is too complex for linear statistical models to capture
adequately.

\subsection{The Dual-Temporal Structure of Booking Data}

A key insight that motivates this work is that airline booking data
carries two distinct and complementary temporal signals, which prior
studies have consistently treated in isolation. The first is the
horizontal sequence: the day-by-day accumulation of bookings for a
single flight as its departure date approaches. This sequence captures
the intra-flight booking momentum, how quickly seats are filling up
relative to historical norms for that flight. The second is the
vertical sequence: the booking status of the same route at the same
number of days before departure, observed across multiple historical
flights. This sequence captures the inter-flight seasonality; the
structural demand patterns that recur across weeks, months, and
seasons for a given route.

Pan~\emph{et~al.}~\cite{pan2018} were among the first to distinguish
these two temporal dimensions in the context of LSTM-based airline
demand forecasting, proposing separate single-stream models for each.
Later, He~\emph{et~al.}~\cite{he2023} refined this approach through
LSTM-T and LSTM-P architectures that separately processed temporal and
positional sequences. Both contributions provided meaningful advances,
but neither integrated the horizontal and vertical streams into a
unified model. A flight might be filling slowly today; a horizontal
signal suggesting low demand, yet historical data for that same route
might consistently show a late-booking surge. A model that sees only
one of these signals cannot reason about the other, and the
information lost by this separation is not marginal.

\subsection{The Capacity Robustness Problem}

A second limitation of the existing literature is the near-universal
choice of absolute passenger count as the forecasting target. While
this is a natural and interpretable metric, it is not operational
friendly. It's common in airlines to change equipment; substituting
one aircraft type for another due to maintenance requirements,
schedule disruptions, or yield management considerations. When a
Boeing~737 with 162 seats is replaced by a Boeing~777 with 419 seats
on the same route, a forecast of, for example, 150 passengers carry
fundamentally different operational meaning. On the narrow-body
aircraft it signals a near-full flight requiring conservative pricing.
On the wide-body it signals a low-load flight requiring stimulation.
A model trained to predict absolute counts provides no mechanism to
distinguish these scenarios.

Passenger Load Factor (PLF), defined as the ratio of booked seats to
total available seat capacity, normalizes the forecasting target
against aircraft capacity and therefore remains interpretable and
consistent regardless of which aircraft operates a given flight.
Predicting PLF instead of absolute passenger counts does not reduce
the model's practical utility, as passenger counts can be deduced from
the predicted PLF and aircraft capacity. This significantly improves
the operational generalizability of the forecast.

\subsection{Research Gaps and Contributions}

The foregoing discussion identifies three interconnected gaps in the
current literature. First, no existing LSTM-based model simultaneously
utilizes both horizontal and vertical booking sequences within a
unified dual-stream architecture. Second, the use of PLF as a
forecasting target, despite its operational advantages, remains
largely unexplored in deep learning approaches to airline demand
prediction. Third, existing studies lack a structured, empirically
justified feature selection pipeline tailored to the high-dimensional
and temporally heterogeneous nature of airline reservation data.

This study addresses all three gaps through the following
contributions:

First, a dual-stream LSTM is proposed that processes
horizontal and vertical booking sequences through parallel LSTM 
branches, whose outputs are combined via a attention based fusion mechanism.
This architecture is the first, to the authors' knowledge, to joint
model intra-flight and inter-flight booking dynamics within a single
end-to-end trainable network.

Second, a novel offset-based algorithm for constructing vertical
sequences is introduced, which identifies historical flight records
sharing the same route and the same days-before-departure value. This
construction method captures inter-flight booking behavior at
structurally equivalent temporal positions, rather than relying on
simple calendar-based alignment.

Third, PLF is adopted as the forecasting target throughout, and its
operational advantages over absolute passenger counts are demonstrated
empirically across multiple aircraft types present in the dataset.

Fourth, a seven-stage feature selection pipeline is developed
combining Pearson correlation, mutual information ranking, random
forest importance, sequential forward selection, variance inflation
factor analysis, and feature deduplication. This pipeline reduced 39
engineered features to 17 optimally informative inputs, separately
optimized for horizontal and vertical streams.

Fifth, six dual-stream architectural variants are systematically
evaluated, incorporating self-attention, cross-attention, hybrid
attention, gated fusion, and residual fusion mechanisms. This provides
a structured comparative analysis of how different integration
strategies affect forecasting performance across the full range of
operationally relevant flight types.

Experiments are conducted on real-world reservation data from the
national carrier of Bangladesh, for the tenure 2023 to 2024, and
validated across four flight category pairs: domestic versus
international, direct versus transit, high versus low frequency, and
short versus long haul. To the best of our knowledge, this is the first study to leverage both intra-flight and inter-flight booking information for short-term airline passenger load forecasting.

\subsection{Paper Organisation}

The remainder of this paper is organised as follows. Section~\ref{sec:lit}
reviews the relevant literature. Section~\ref{sec:method} describes
the dataset, feature engineering, sequence construction, model
architectures, and hyperparameter optimisation. Section~\ref{sec:exp}
details the experimental setup and evaluation protocol.
Section~\ref{sec:results} reports and discusses the results.
Section~\ref{sec:conclusion} concludes the paper and identifies
directions for future research.

% ============================================================
%  II. LITERATURE REVIEW
% ============================================================
\section{Literature Review}
\label{sec:lit}

\subsection{Statistical and Econometric Approaches}

Early airline demand forecasting relied primarily on statistical
time-series methods. Autoregressive Integrated Moving Average (ARIMA)
models and their seasonal variant (SARIMA) were widely applied to
predict aggregate passenger volumes at flight route levels.
Nieto and Carmona-Ben\'itez~\cite{nieto2018} combined ARIMA with
GARCH and bootstrap resampling to manage varying volatility in monthly
airline traffic data. Later, Carmona-Ben\'itez and
Nieto~\cite{carmona2020} extended this work using a SARIMA-based grey
forecasting model to improve accuracy on seasonal series.
Wu~\emph{et~al.}~\cite{wu2021} applied a two-phase learning model to
forecast air passenger traffic flows at airport level.
Grimme~\emph{et~al.}~\cite{grimme2020} and Dey
Tirtha~\emph{et~al.}~\cite{dey2022} analyzed total
origin-destination demand using airport-level data, focusing on how
the COVID-19 pandemic affected passenger numbers.

Econometric approaches extended this work by incorporating
macroeconomic predictors. Suryan~\cite{suryan2017} examined the role
of GDP, income levels, and population in forecasting air passenger
demand in Indonesia, while Bastola~\cite{bastola2017} conducted
similar analysis for Nepal. Haensel~\emph{et~al.}~\cite{haensel2011}
used airline reservation data to estimate unconstrained customer
choice set demand. They noted the censored nature of booking
observations. Dynamic booking forecasting frameworks by van
Ostaijen~\emph{et~al.}~\cite{vanostaijen} and
Marques~\cite{marques2016} demonstrated the predictive value of
reservation data over aggregate statistics, particularly for
short-term operational decisions.

However, these methods assume either linearity or stationarity in
demand behavior, and most operate at monthly or quarterly temporal
resolution. This granularity is insufficient for flight-level revenue
management, which requires daily or sub-daily forecasts tied to the
evolving booking curve.

\subsection{Machine Learning Approaches}

Machine learning methods addressed the non-linearity limitation of
statistical models. Artificial neural networks~\cite{firat2021},
decision trees, and k-means clustering~\cite{chen2020} were applied
to aviation demand with improved flexibility.
Hopman~\emph{et~al.}~\cite{hopman2021} proposed a machine learning
approach to itinerary-level booking prediction within competitive
airline environments, focusing on estimating the likelihood of
individual passengers selecting particular itineraries.
Firat~\emph{et~al.}~\cite{firat2021} applied several ML algorithms to
forecast air travel demand across selected destinations.
Chen~\emph{et~al.}~\cite{chen2020} used k-means clustering combined
with decision tree classification to identify critical macroeconomic
factors driving air traffic volume.

While ML models improved on the linearity constraints of earlier
methods, they presented their own limitations. Most treat booking
observations as independent samples and do not exploit the sequential
structure of the booking curve. Ghandeharioun~\emph{et~al.}~\cite{ghandeharioun2023}
observed that shallow ML models generally failed to capture temporal
dependencies in short-term passenger demand, particularly when booking
behavior exhibited irregular seasonality or late-surge patterns.

\subsection{Deep Learning and Single-Stream LSTM Approaches}

To address the vanishing gradient problem inherent in standard
recurrent neural networks~\cite{rumelhart1987}, Hochreiter and
Schmidhuber~\cite{hochreiter1997} introduced Long Short-Term Memory
(LSTM) networks. This capability of LSTM makes it highly effective for
analyzing sequential data like airline reservation data with
long-range temporal dependencies. Do~\emph{et~al.}~\cite{do2020}
demonstrated that LSTM outperformed SARIMA in forecasting weekly and
monthly passenger counts at Incheon Airport, establishing the use of
deep learning in aviation time-series applications.
Kanavos~\emph{et~al.}~\cite{kanavos2021} developed ARIMA, SARIMA, and
Deep Learning Neural Network (DLNN) models for aviation demand. Which
demonstrated through qualitative comparison that DLNN provided more
robust results. Ghandeharioun~\emph{et~al.}~\cite{ghandeharioun2023}
further confirmed the superiority of LSTM over conventional ML models
for short-term passenger demand prediction. Furthermore,
Iacus~\emph{et~al.}~\cite{iacus2020} demonstrated the adaptability of
these models by projecting air passenger traffic disruption during the
COVID-19 pandemic.

The two most directly relevant single-stream LSTM studies are
Pan~\emph{et~al.}~\cite{pan2018} and He~\emph{et~al.}~\cite{he2023}.
Pan~\emph{et~al.}~\cite{pan2018} were among the first to apply LSTM
to flight-level booking data, formally distinguishing between the
horizontal sequence; daily booking accumulation for a single flight
and the vertical sequence; booking records for the same route at fixed
days-before-departure across historical flights. They trained separate
models for each sequence type but did not combine them.
He~\emph{et~al.}~\cite{he2023} refined this framework and obtained
improved accuracy through LSTM-T and LSTM-P architectures that
processed temporal and positional booking sequences independently.
Both studies represent meaningful progress, but neither integrates the
two temporal dimensions within a unified architecture, and both
predict absolute passenger counts rather than a capacity-normalized
target.

\subsection{Dual-Stream LSTM Architectures}

Dual-stream LSTM architectures, where two parallel LSTM branches
process distinct input sequences and their outputs are fused, have
been explored in domains outside aviation.
Islam~\emph{et~al.}~\cite{islam2021} proposed an efficient dual-stream
deep learning architecture combining Separable Convolutional LSTM and
pre-trained MobileNet for violence detection in video. They
demonstrated that parallel temporal streams can capture complementary
representational aspects of the same event.
Gu~\emph{et~al.}~\cite{gu2022} applied a Dual Input Attention LSTM
(DIA-LSTM) to agricultural commodity price forecasting, where one
input stream carried historical prices and the other carried related
external factors. Self-attention was applied to each stream
independently before concatenation, which enabled the model to weight
each time step's contribution dynamically.
Peng~\emph{et~al.}~\cite{peng2022} predicted the remaining useful life
of industrial equipment using a dual-channel LSTM. They processed
original sensor features in one branch and their first-order
differences, approximating rate-of-change signals in the other,
fusing the outputs elementwise.

These studies share an important structural commonality with the
present work: each process two related but distinct temporal signals
through separate LSTM branches and combines their representations to
improve prediction accuracy. However, none is applied to airline
booking data, and none addresses the specific challenge of fusing
intra-flight and inter-flight temporal patterns for load factor
forecasting.

\subsection{Forecasting Target: Passenger Count Versus PLF}

Most demand forecasting studies in aviation predict absolute passenger
counts, either at the flight level~\cite{pan2018,he2023} or at
aggregate airport and origin-destination
levels~\cite{nieto2018,carmona2020,wu2021}. This approach aligns with
traditional revenue management, which focuses on available seats.
However, this method introduces a dependency on aircraft capacity. If
the aircraft type changes, the model's prediction may no longer be
valid, limiting the model's applicability.

Passenger Load Factor (PLF), defined as the ratio of revenue passenger
kilometers to available seat kilometers, normalizes demand against
capacity and therefore remains stable across aircraft type changes.
Despite its operational advantages, PLF has received limited attention
as a forecasting target in deep learning research.
Kevser~\emph{et~al.}~\cite{simsek2024} is the only known prior work
to use PLF as the direct target variable, where fractional calculus
was applied within a deep assessment methodology to forecast PLF from
reservation data. To the authors' knowledge, no LSTM-based model has
targeted PLF using dual-temporal booking sequences.

\section{Methodology}
\label{sec:method}

\subsection{Dataset Description}
The dataset used in this study was constructed by integrating three
sources: flight-level reservation records from the national carrier of
Bangladesh, airports data, and the national holiday calendar of
Bangladesh. The reservation data covers a two-year period (2023--2024)
and contains daily snapshots of booking status for each scheduled
flight. Which were recorded each day from 30 days before departure
through to the departure date itself. Raw leg-wise flight records were
aggregated to flight-date-level records and then filtered to individual
route-level datasets (${\sim}12{,}806$ records per route after
cleaning) to construct model inputs. The dataset incorporates the
entirety of the carrier's operational flight spectrum. Which includes
routes classified by geographical reach (domestic and international),
service structure (direct and transit), distance (short-haul to
long-haul), and operational tempo (high and low weekly frequencies).
Airports data contained IATA codes, geographic coordinates, and route
distance information. Holiday data introduced binary calendar
indicators for public and national holidays in Bangladesh.

\subsection{Problem Formulation}
Rather than predicting absolute passenger counts, this study adopts Passenger Load Factor (PLF) as the forecasting target. PLF is defined as:
\begin{align}
\text{PLF} &= \frac{\text{RPK}}{\text{ASK}}
             = \frac{\text{Booked Seats} \times \text{Distance}}
                    {\text{Total Seats} \times \text{Distance}}
             = \frac{\text{Booked Seats}}{\text{Total Seats}},
\label{eq:plf}
\end{align}

\begin{figure}[!t]
\centering

\includegraphics[width=\columnwidth]{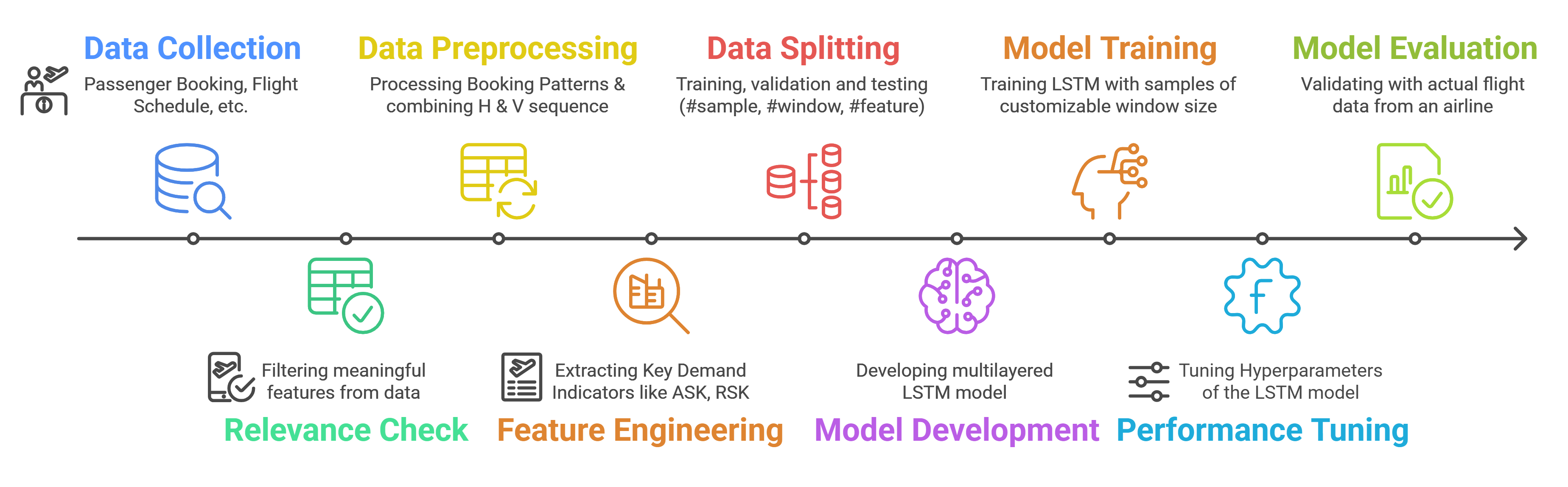}
\caption{Pipeline of proposed approach.}
\label{fig:pipeline}
\end{figure}

where RPK denotes Revenue Passenger Kilometers and ASK denotes Available Seat Kilometers. Here, PLF gets simplified to the ratio of booked seats to total seat capacity, as distance cancels in the route-level computation. This normalization decouples the forecast from absolute aircraft capacity. This ensures that predictions remain operationally valid when equipment changes alter the total number of available seats on a given flight. The original passenger count can be recovered from the PLF forecast when aircraft capacity is known. This preserves the model's practical application in revenue management. Fig.~\ref{fig:pipeline} illustrates the end-to-end pipeline of the proposed approach.

\subsection{Feature Engineering}

A total of 39 features were engineered from the three data sources.
The engineering process proceeded through five stages: reading and
cleaning daily reservation files, geospatial enrichment via airport
master data to compute route distances, calculation of operational
KPIs (ASK, RPK, PLF), integration of calendar features, and
construction of rolling and historical demand statistics.

Calendar features were encoded using sine-cosine transformations to
preserve their cyclic nature. For example, day-of-week was encoded as:
\begin{equation}
\text{day\_sin} = \sin\!\left(\frac{2\pi \cdot \text{day}}{7}\right),\quad
\text{day\_cos} = \cos\!\left(\frac{2\pi \cdot \text{day}}{7}\right).
\label{eq:sincos}
\end{equation}
The same transformation was applied to month, day-of-month, and
week-of-year features for both the flight date and the booking record
date. Two demand-related rolling statistics were computed:
\texttt{rolling\_avg\_plf}, the moving average of PLF across recent
booking records, and \texttt{plf\_historical}, the average PLF for the
same flight and days-before-departure position across historical
flights. These two features serve as direct proxies for short-term
booking momentum and long-term route-level demand baseline,
respectively. Features were categorised as horizontal (H), vertical
(V), or shared (H+V) based on their temporal semantics, as detailed in
Table~\ref{tab:features}.

%\subsection{Feature Selection Pipeline}

Airline reservation data is complex, with many features that overlap,
are highly correlated, and have varying degrees of relevance to the
prediction target. To address this, a seven-stage selection pipeline
was developed, applied independently to the horizontal and vertical
feature pools to maintain their unique time-related characteristics.

\emph{Stage~1 --- Redundancy elimination via Pearson correlation:}
Pairwise Pearson correlation coefficients were computed for all
candidate features. Where two features exceeded a correlation
threshold of 0.90, the less domain-relevant feature was removed.

\emph{Stage~2 --- Non-linear relevance ranking via Mutual Information
(MI):} MI scores were computed between each remaining feature and the
PLF target variable. Unlike Pearson correlation, MI captures
non-linear dependencies and imposes no distributional assumptions.

\emph{Stage~3 --- Supervised importance estimation via Random Forest
(RF):} A Random Forest Regressor was trained independently on each
temporal stream. Feature importance scores were derived from mean
impurity reduction across trees. Horizontal features showed strong
concentration of importance around \texttt{total\_RPK} and
\texttt{rolling\_avg\_plf}, while vertical features exhibited more
distributed importance across holiday indicators and historical PLF
statistics.

\emph{Stage~4 --- Wrapper-based selection via Sequential Forward
Selection (SFS):} SFS iteratively added features to a Ridge
Regression estimator, evaluating predictive gain at each step. SFS
yielded a compact subset of 20 features that collectively maximised
predictive value. Overlap analysis between MI, RF, and SFS showed
that \texttt{total\_RPK} and \texttt{rolling\_avg\_plf} were
consistently selected by all three methods.

\emph{Stage~5 --- Multicollinearity assessment via Variance Inflation
Factor (VIF):} VIF was computed for all SFS-selected features. In
LSTM architectures, coefficient instability from multicollinearity
does not apply; the network learns non-linear transformations of
inputs. Accordingly, features with elevated VIF values were not
automatically excluded. Features exhibiting extreme multicollinearity
(\texttt{rolling\_avg\_plf\_H}, VIF\,$\approx$\,1581;
\texttt{total\_RPK\_H}, VIF\,$\approx$\,1146) were retained on the
basis of their strong and consistent signal across Stages~2, 3, and~4,
following established practice in neural network-based feature
retention.

\emph{Stage~6 --- Deduplication:} Features selected across MI, RF,
and SFS were mapped back to their base feature names by removing
temporal suffixes (e.g., \_H3, \_V2). Repeated derivations of the
same base feature were consolidated to retain the most frequently
selected temporal variant.

\emph{Stage~7 --- Final stream-specific selection:} From the
deduplicated pool, the top-ranked features were selected separately
for horizontal and vertical inputs.

\begin{table}[!t]
\caption{Final Selected Features for Horizontal (H) and Vertical (V) Sequence}
\label{tab:features}
\centering
\renewcommand{\arraystretch}{1.2}
\begin{tabularx}{\columnwidth}{X c c X}
\toprule
\textbf{Feature} & \textbf{H} & \textbf{V} & \textbf{Description} \\
\midrule
\texttt{plf} & \checkmark & \checkmark & Passenger Load Factor (target proxy in sequence) \\
\texttt{total\_RPK} & \checkmark & \checkmark & Revenue Passenger Kilometers \\
\texttt{rolling\_avg\_plf} & \checkmark & \checkmark & Rolling average PLF across booking window \\
\texttt{plf\_historical} & \checkmark & \checkmark & Historical average PLF at same days-before-departure \\
\texttt{days\_before\_departure} & \checkmark & \checkmark & Days between record date and flight date \\
\texttt{record\_date\_day} & \checkmark &  & Day of month of booking observation \\
\texttt{record\_date\_day\_of\_year} & \checkmark &  & Day of year of booking observation \\
\texttt{flight\_date\_is\_holiday} & \checkmark & \checkmark & Binary flag: flight date falls on a public holiday \\
\texttt{flight\_date\_day} &  & \checkmark & Day of month of flight date \\
\texttt{flight\_date\_week} &  & \checkmark & Week number of flight date \\
\texttt{flight\_date\_day\_of\_year} &  & \checkmark & Day of year of flight date \\
\texttt{flight\_date\_is\_weekend} &  & \checkmark & Binary flag: flight date falls on a weekend \\
\bottomrule
\end{tabularx}
\end{table}

\begin{figure}[!t]
\centering
\includegraphics[width=0.72\columnwidth]{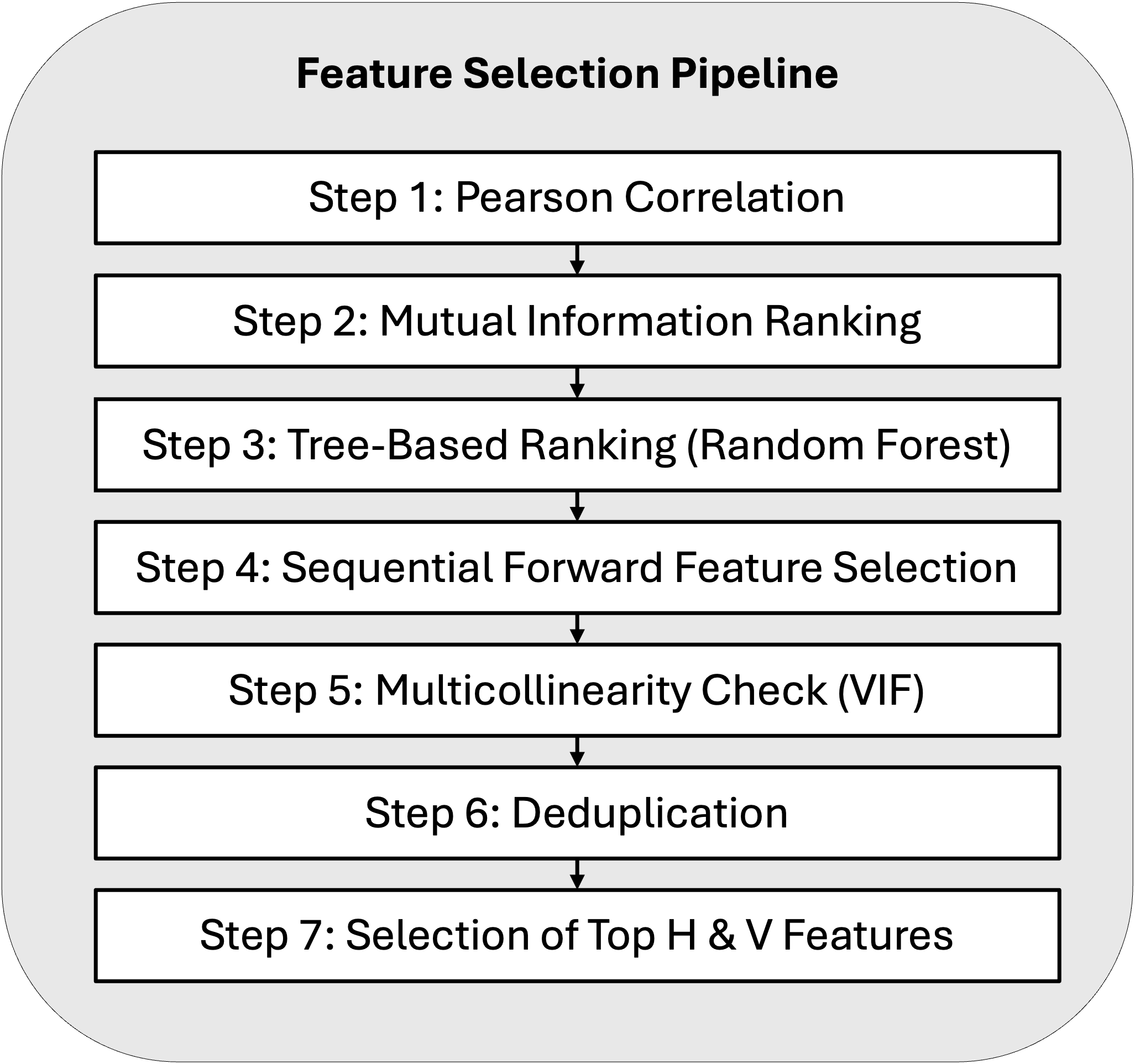}
\caption{Feature selection pipeline.}
\label{fig:featsel}
\end{figure}

Table~\ref{tab:features} contains the finally selected features
comprising 8 horizontal features and 9 vertical features (17 total).
The horizontal set is dominated by booking-momentum and temporal
proximity features, while the vertical set emphasises historical
seasonality, holiday proximity, and route-level demand averages.
Fig.~\ref{fig:featsel} summarises the seven-stage feature selection
pipeline.

\subsection{Sequence Construction}

The model requires two structured input sequences for each prediction
instance. Both are constructed from the same underlying dataset but
capture different temporal dimensions of booking behaviour.

\emph{Horizontal sequence:} For a given flight on a given flight date,
the horizontal sequence captures the booking progression over the $h$
days immediately preceding the prediction point. Each element of the
sequence is a feature vector recorded on a specific
\texttt{days\_before\_departure} value. Which is ordered
chronologically from oldest to most recent. For window size $h=3$,
the sequence contains three consecutive booking snapshots at
days-before-departure positions $[d,\,d{+}1,\,d{+}2]$, where $d$ is
the current prediction day. This sequence enables the model to learn
intra-flight booking momentum.

\emph{Vertical sequence:} The vertical sequence captures inter-flight
booking patterns; how flights on the same route historically behaved
at the same days-before-departure position. For a given flight
departing on date $t$ and recorded at $d$ days before departure, the
vertical sequence is constructed by locating the $v$ most recent
historical flights on the same route that also have records at exactly
$d$ days before departure. Each element of the sequence is the feature
vector for one such historical flight, ordered from oldest to most
recent. This offset-based construction is the key methodological distinction
from simple calendar-based alignment. Rather than selecting historical
flights by calendar proximity alone, the vertical sequence ensures
that each historical record occupies the same structural position on
the booking curve i.e., the same days-before-departure value as the
current prediction instance. This preserves the temporal semantics of
the vertical signal across flights with varying departure dates and
booking patterns. Two approaches to vertical sequence generation were evaluated. Approach~1 selects the $v$ most recent historical flights at the same days-before-departure offset. Approach~2 applies an additional
temporal stride between historical records to increase the span of
booking history captured. Approach~1 produced consistently better
performance across all model variants and was adopted for the final
architecture. Both sequences use a window size of 3 ($H{=}3$, $V{=}3$), determined through a systematic grid search over candidate sizes ranging from 3 to 18, evaluated on validation loss across all four base model
architectures. The symmetric window ($H{=}V{=}3$) outperformed all
asymmetric configurations tested.

\subsection{Model Architectures}

Four baseline architectures were developed to systematically isolate the
contribution of each design decision, followed by six dual-stream
variants that evaluate fusion mechanisms.

\emph{Single-Stream Baselines:} Three single-stream configurations
serve as baselines. SLSTM-H processes only the horizontal sequence
through a stacked LSTM network. SLSTM-V processes only the vertical
sequence through an analogous architecture. SLSTM-C integrates both
input sequences by concatenating them into a unified feature vector,
which is then processed by a single LSTM network.

\emph{Dual-Long Short-Term Memory Based Architecture (DLSTM):} The dual-stream base
model processes horizontal and vertical sequences through two
independent LSTM branches in parallel. The horizontal branch takes an
input of shape $(3, 8)$; the vertical branch takes input of shape
$(3, 9)$. The outputs of both branches are concatenated and passed
through a dense layer to produce the PLF prediction.

\emph{Attention-Based Variants:} Three attention variants were
developed on the DLSTM base. DLSTM-SA applies self-attention to the
output sequence of each LSTM branch independently before
concatenation. DLSTM-CA applies cross-attention between the two
branches: the horizontal stream attends to the vertical, and vice
versa. Long Short-Term Memory Based hybrid attention (DLSTM-HA) combines both mechanisms sequentially: self-attention is first applied within each branch, followed by cross-attention
between branches. This hybrid arrangement captures both intra-stream
temporal salience and inter-stream contextual alignment within a
single forward pass.

\begin{figure}[!t]
\centering
\includegraphics[width=\columnwidth]{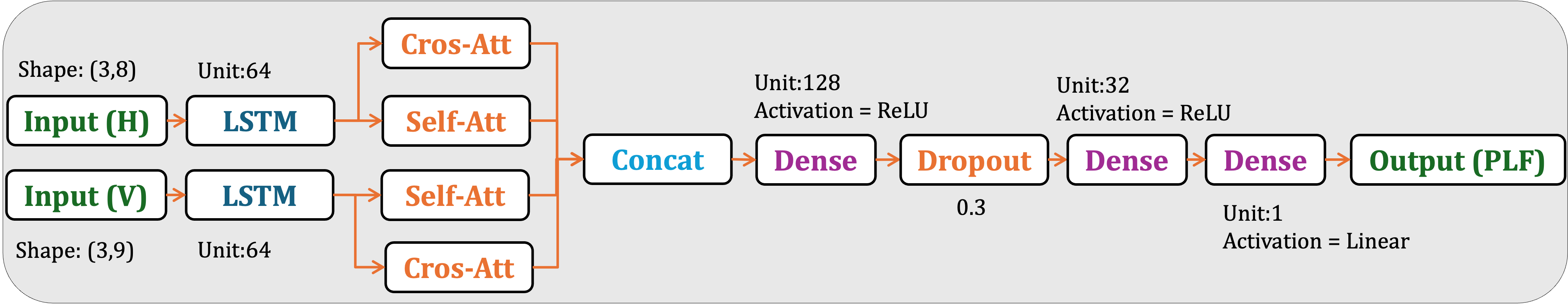}
\caption{DLSTM-HA model architecture.}
\label{fig:architecture}
\end{figure}

\emph{Fusion-Based Variants:} Two additional variants explore
alternative fusion strategies applied after hybrid attention.
DLSTM-HARF replaces the concatenation fusion in DLSTM-HA with
residual fusion: the attended representation of each branch is added
element-wise to the original LSTM output of that branch before
combination, with the intuition that the original LSTM output may
carry information that attention selectively suppresses. DLSTM-HAGF
applies gated fusion, where a learned gate vector determines the
contribution weight of each stream's representation before
combination. Fig.~\ref{fig:architecture} presents the architecture of
the proposed Dual Long Short-Term Memory Based hybrid attention (DLSTM-HA) model.

\subsection{Hyperparameter Optimisation}

\begin{table}[!t]
\caption{Optimal Hyperparameter Configurations for Base Models}
\label{tab:hyperparam}
\centering
\renewcommand{\arraystretch}{1.2}
\begin{tabular}{lcccccc}
\toprule
\textbf{Model} & \textbf{LSTM} & \textbf{Lyr} & \textbf{Drop} & \textbf{Dense} & \textbf{LR} & \textbf{Opt.} \\
\midrule
SLSTM-H & 96          & 2 & 0.1 & 32 & 0.004337 & Adam    \\
SLSTM-V & 128         & 1 & 0.1 & 32 & 0.001318 & RMSprop \\
SLSTM-C & 128         & 1 & 0.1 & 32 & 0.001318 & RMSprop \\
DLSTM   & {[}96,32{]} & 2 & 0.1 & 32 & 0.004337 & Adam    \\
\bottomrule
\end{tabular}
\end{table}

Hyperparameters for all model variants were tuned using Keras Tuner
with a Bayesian optimisation strategy. The search space covered LSTM
units $\{32, 64, 96, 128\}$, number of LSTM layers $\{1, 2\}$,
dropout rate $\{0.1, 0.2, 0.3, 0.4, 0.5\}$, dense layer units
$\{32, 64, 128\}$, learning rate ($1{\times}10^{-4}$ to
$1{\times}10^{-2}$), and optimiser \{Adam, RMSprop\}. Tuning was
conducted independently for each model variant. Early stopping with
patience of 5 epochs and ReduceLROnPlateau with a reduction factor of
0.5 were applied during both tuning and final training. The optimal
configurations derived for the four base models are summarised in
Table~\ref{tab:hyperparam}. Attention and fusion variants inherited
the DLSTM base configuration with LSTM units set to 64 per branch,
dropout of 0.3, and a two-layer dense head of units $\{128, 32\}$,
based on tuned configurations validated during the hyperparameter
search.

% ============================================================
%  IV. EXPERIMENTAL SETUP
% ============================================================
\section{Experimental Setup and Evaluation}
\label{sec:exp}

\subsection{Data Partitioning}

The dataset was divided chronologically into training (70\%),
validation (15\%), and test (15\%) subsets to preserve the temporal
order of records. To prevent leakage of future booking information
into the training set, random splitting was avoided. All input
features were standardised using a StandardScaler fitted on the
training dataset only. Then applied to validation and test datasets
without re-fitting to prevent data leakage through the normalisation
step.

\subsection{Training Configuration}

All LSTM models, single-stream and dual-stream were trained under a
consistent configuration to ensure fair comparison across variants.
The Adam optimiser was used with Mean Squared Error (MSE) as the loss
function. A batch size of 32 was fixed across all experiments.
Training ran for a maximum of 100 epochs, with two callbacks applied
to manage convergence. Early stopping monitored validation loss with a
patience of 5 epochs, halting training when no improvement was
observed and restoring the best weights. ReduceLROnPlateau reduced the
learning rate by half when validation loss plateaued for 3 consecutive
epochs, with a minimum learning rate of $1{\times}10^{-5}$. These
settings were applied uniformly across all models.

\subsection{Evaluation Metrics}

Model performance was assessed using six metrics, each capturing a
different aspect of forecasting error. This breadth of evaluation was
necessary because a singular measure is insufficient for comprehensive
evaluation, particularly given the varying degrees of volatility
observed in PLF across different flight types. Mean Absolute Error
(MAE) measures the average magnitude of prediction errors in the same
unit as PLF. Mean Absolute Percentage Error (MAPE) expresses errors
as a proportion of the actual value, enabling performance comparison
across different flights. Mean Squared Error (MSE) and Root Mean
Squared Error (RMSE) penalise larger errors more heavily. Mean
Absolute Scaled Error (MASE) normalises the absolute error against the
error of a na\"ive benchmark forecast; MASE\,$<$\,1 indicates the
model outperforms the na\"ive baseline. The Coefficient of
Determination ($R^{2}$) measures the proportion of PLF variance
explained by the model.

\subsection{Baseline Models}

To contextualise the performance of the proposed dual-stream
architecture, three categories of comparison were conducted. First,
four standard non-sequential machine learning models were used as
benchmarks: Linear Regression (LR), Random Forest (RF), XGBoost, and
LightGBM. These models received the same features as the LSTM models
but processed them as flat feature vectors without temporal structure.
Second, the two most directly comparable prior studies,
Pan~\emph{et~al.}~\cite{pan2018} and He~\emph{et~al.}~\cite{he2023}
were reimplemented and evaluated on the same dataset under identical
conditions, in both their horizontal-only (H) and vertical-only (V)
configurations. Third, three prior dual-LSTM architectures, Islam~\emph{et~al.}~\cite{islam2021},
Gu~\emph{et~al.}~\cite{gu2022} and Peng~\emph{et~al.}~\cite{peng2022}
were similarly reimplemented on the same dataset.

% ============================================================
%  V. RESULTS AND DISCUSSION
% ============================================================
\section{Results and Discussion}
\label{sec:results}

%\subsection{Ablation Study: Contribution of Each Temporal Stream}

\begin{table}[!t]
\caption{Ablation Results --- Symmetric Window ($H{=}3$, $V{=}3$)}
\label{tab:ablation}
\centering
\renewcommand{\arraystretch}{1.2}
\begin{tabular}{lcccc}
\toprule
\textbf{Metric} & \textbf{SLSTM-H} & \textbf{SLSTM-V} & \textbf{SLSTM-C} & \textbf{DLSTM} \\
\midrule
MAE     & $2.96{\pm}0.18$  & $11.76{\pm}0.26$ & $2.93{\pm}0.04$ & $3.00{\pm}0.12$ \\
MAPE    & $0.10{\pm}0.01$  & $0.41{\pm}0.02$  & $0.11{\pm}0.00$ & $0.10{\pm}0.01$ \\
MSE     & $26.82{\pm}0.62$ & $231.97{\pm}4.08$& $26.60{\pm}0.39$& $27.90{\pm}0.73$\\
RMSE    & $5.18{\pm}0.06$  & $15.23{\pm}0.13$ & $5.16{\pm}0.04$ & $5.28{\pm}0.07$ \\
MASE    & $0.59{\pm}0.04$  & $2.33{\pm}0.05$  & $0.58{\pm}0.01$ & $0.59{\pm}0.02$ \\
$R^{2}$ & $0.95{\pm}0.00$  & $0.56{\pm}0.01$  & $0.95{\pm}0.00$ & $0.95{\pm}0.00$ \\
\bottomrule
\end{tabular}
\end{table}

To understand what each input stream contributes to forecasting
accuracy, single-stream and dual-stream models were first evaluated in
isolation using a symmetric window size of H=3, V=3. The results are
presented in Table~\ref{tab:ablation}. 

The vertical-only model (SLSTM-V) performed substantially worse than all other configurations, with MAE\,$=\,11.76\,{\pm}\,0.26$ and $R^{2}\,=\,0.56\,{\pm}\,0.01$. The vertical sequence captures booking patterns at a fixed days-before-departure position across historical flights. This helps in understanding seasonal demand, but it provides almost no information about how the current flight is filling up. Without the horizontal booking progression, the model is
essentially working from historical averages alone. 

The horizontal-only model (SLSTM-H) performed considerably better,
achieving MAE\,$=\,2.96\,{\pm}\,0.18$ and
$R^{2}\,=\,0.95\,{\pm}\,0.00$. This confirms intra-flight booking
accumulation carries the dominant predictive information for short-term
PLF. However, as the model sees only the current flight's trajectory,
therefore cannot assess whether this trajectory is normal or unusual
relative to how similar flights historically behaved at the same stage.

The combined single-stream model (SLSTM-C), which concatenates
horizontal and vertical features into one input vector, achieved
MAE\,$=\,2.93\,{\pm}\,0.04$. The base dual-stream model (DLSTM),
processing sequences through separate branches, achieved
MAE\,$=\,3.00\,{\pm}\,0.12$. The dual-stream architecture's advantage
becomes apparent only when attention mechanisms are added, as
discussed in Section~\ref{subsec:attn}.

\begin{table*}[!t]
\caption{Performance of Dual-Stream Variants}
\label{tab:dualvariants}
\centering
\renewcommand{\arraystretch}{1.2}
\begin{tabular}{lcccccc}
\toprule
\textbf{Metric} & \textbf{DLSTM-SA} & \textbf{DLSTM-CA} & \textbf{DLSTM-HA} & \textbf{DLSTM-HAGF} & \textbf{DLSTM-HARF} & \textbf{DLSTM-HAGFRF} \\
\midrule
MAE     & $2.9628{\pm}0.0225$ & $2.8406{\pm}0.0676$ & $\mathbf{2.8167{\pm}0.0379}$ & $2.8496{\pm}0.0348$ & $2.9275{\pm}0.0207$ & $2.9211{\pm}0.0185$ \\
MAPE    & $0.0996{\pm}0.0029$ & $0.0867{\pm}0.0005$ & $0.0874{\pm}0.0044$          & $0.0948{\pm}0.0018$ & $0.0939{\pm}0.0049$ & $0.0958{\pm}0.0008$ \\
MSE     & $27.8678{\pm}0.1743$& $26.9655{\pm}1.1890$& $26.5314{\pm}0.2085$         & $26.2177{\pm}0.1294$& $27.0326{\pm}0.7318$& $26.7495{\pm}0.3268$\\
RMSE    & $5.2790{\pm}0.0165$ & $5.1916{\pm}0.1137$ & $5.1508{\pm}0.0202$          & $5.1203{\pm}0.0126$ & $5.1988{\pm}0.0700$ & $5.1719{\pm}0.0316$ \\
MASE    & $0.5868{\pm}0.0044$ & $0.5626{\pm}0.0134$ & $\mathbf{0.5579{\pm}0.0075}$ & $0.5644{\pm}0.0069$ & $0.5798{\pm}0.0041$ & $0.5786{\pm}0.0037$ \\
$R^{2}$ & $0.9470{\pm}0.0003$ & $0.9487{\pm}0.0023$ & $0.9495{\pm}0.0004$          & $\mathbf{0.9501{\pm}0.0002}$ & $0.9486{\pm}0.0014$ & $0.9491{\pm}0.0006$ \\
\bottomrule
\end{tabular}
\end{table*}

\subsection{Effect of Attention and Fusion Mechanisms}
\label{subsec:attn}

Six dual-stream variants incorporating different attention and fusion
strategies were evaluated. All variants were evaluated under
Approach~1 vertical sequence generation and symmetric window
($H{=}3$, $V{=}3$). Results are presented in
Table~\ref{tab:dualvariants}.

The self-attention model (DLSTM-SA) improved upon the base DLSTM with
MAE\,$=\,2.9628\,{\pm}\,0.0225$. Cross-attention (DLSTM-CA) brought
further improvement to MAE\,$=\,2.8406\,{\pm}\,0.0676$. This
indicates that enabling direct information exchange between the
horizontal and vertical branches is more valuable than intra-stream
weighting alone.

The hybrid attention model (DLSTM-HA) achieved the best MAE of
$2.8167\,{\pm}\,0.0379$ and $R^{2}$ of $0.9495\,{\pm}\,0.0004$. The
outperformance of DLSTM-HA over DLSTM-CA on MAE suggests that
applying self-attention before cross-attention i.e., effectively
refining each stream's internal representation before inter-stream
fusion leads to a more informative combined representation.

The gated fusion model (DLSTM-HAGF) achieved
MAE\,$=\,2.8496\,{\pm}\,0.0348$ and the highest $R^{2}$ of
$0.9501\,{\pm}\,0.0002$. The most complex variant, DLSTM-HA-GF-RF,
which combines all three fusion strategies, did not improve over
DLSTM-HA or DLSTM-HAGF. Which suggests that stacking multiple fusion
mechanisms introduces redundancy. DLSTM-HA was selected as the
proposed model because of its optimal MAE, which is the most
operationally interpretable metric for airline revenue managers.

\begin{figure}[!t]
\centering
\begin{subfigure}{0.45\textwidth}
  \centering
  \includegraphics[width=\linewidth]{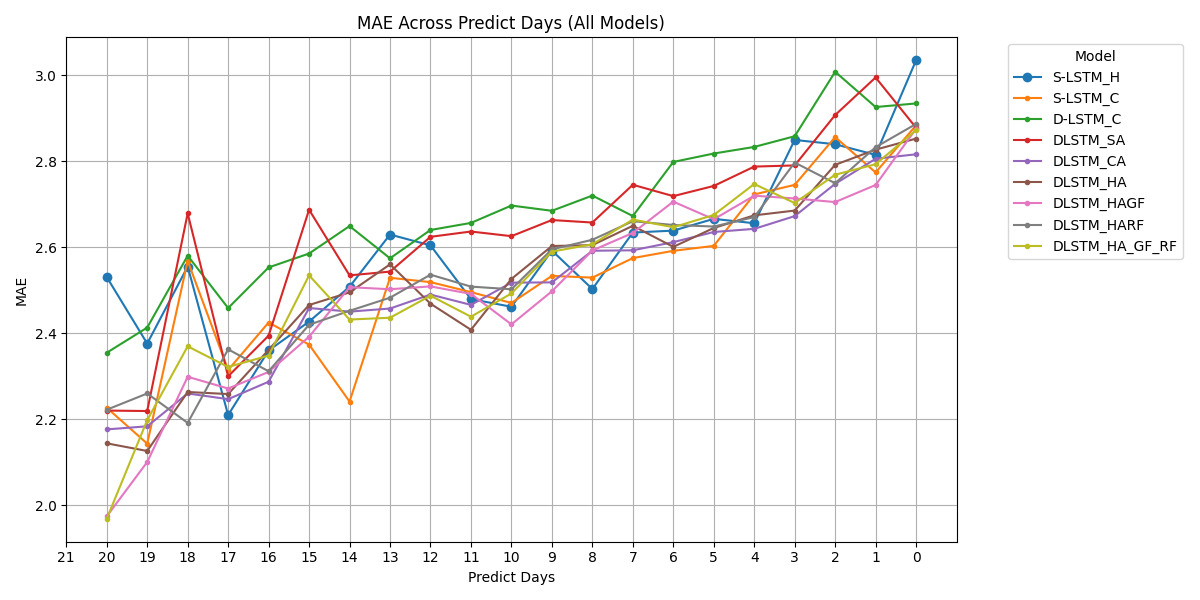}
  \caption{MAE}
\end{subfigure}
\hfill
\begin{subfigure}{0.45\textwidth}
  \centering
  \includegraphics[width=\linewidth]{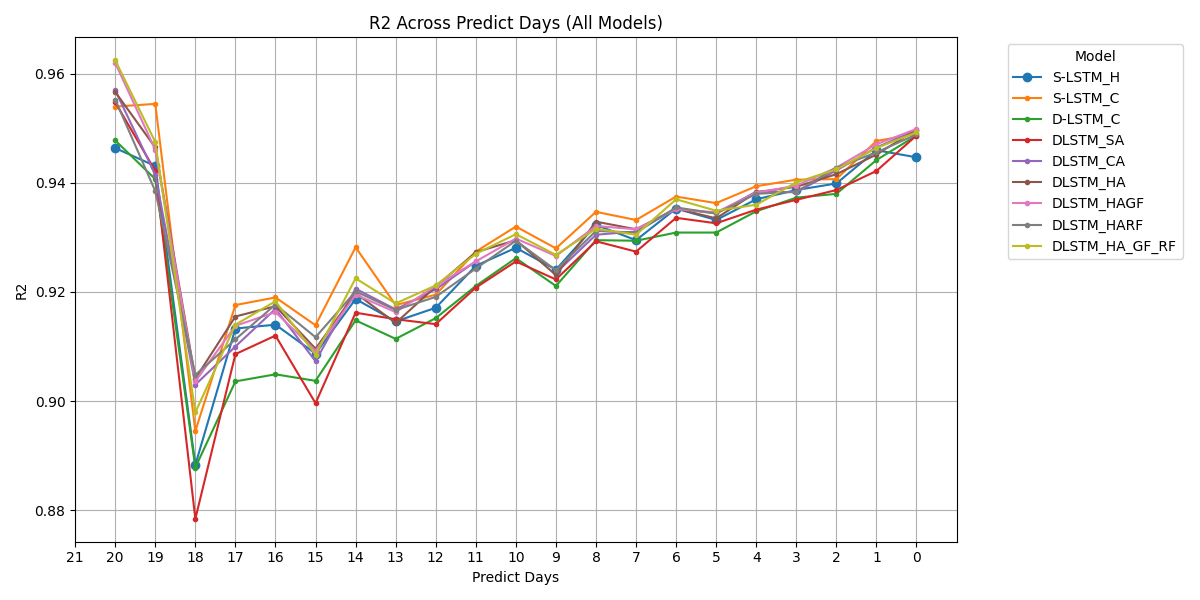}
  \caption{$R^{2}$}
\end{subfigure}
\caption{Trends of MAE (top) and $R^{2}$ (bottom) across dynamic
prediction days (D-21 to D-0).}
\label{fig:horizon}
\end{figure}

\subsection{Dynamic Prediction Horizon Analysis}

One of the practical concerns in airline demand forecasting is how
model performance changes depending on how many days before departure
the prediction is made. To examine this, all model variants were
tested across a dynamic prediction horizon ranging from D-21 to D-0.
Fig.~\ref{fig:horizon} shows MAE and $R^{2}$ values across this
horizon for all models.

In contrast to the expectation of accuracy improving steadily as
departure approaches, the results reveal a more nuanced pattern. From
D-21 to approximately D-8, MAE fluctuates without a clear directional
trend across all models. This variability in the intermediate booking
window reflects the unpredictability in demand from both past flights
and the current flight's accumulation. Beyond D-8, a consistent
increase in MAE is observed across all models as D-0 approaches.
Rather than becoming easier to predict as more booking information
accumulates, flights become harder to predict in the final week before
departure. This pattern reflects the influence of last-minute booking
dynamics: cancellations, upgrades, group bookings, and no-shows, which
are harder to predict despite having complete booking information.

The $R^{2}$ trend follows a similarly non-monotonic path. $R^{2}$
starts moderately high at D-21, drops sharply around D-18 to D-19
(likely when early bookings stop and late bookings have not yet
started), then recovers and stabilises near to the departure date.
Across the full horizon, dual-temporal models consistently
outperformed single-stream baselines. DLSTM-HA maintained the lowest
MAE and highest $R^{2}$ across most prediction days. The vertical-only
model (SLSTM-V) was the most sensitive to horizon length. As past
cross-flight booking patterns become less useful when current booking
conditions differ markedly from the past.

\subsection{Generalisability Across Flight Categories}

\begin{figure*}[!t]
\centering
\begin{subfigure}{0.45\textwidth}
  \centering
  \includegraphics[width=\linewidth]{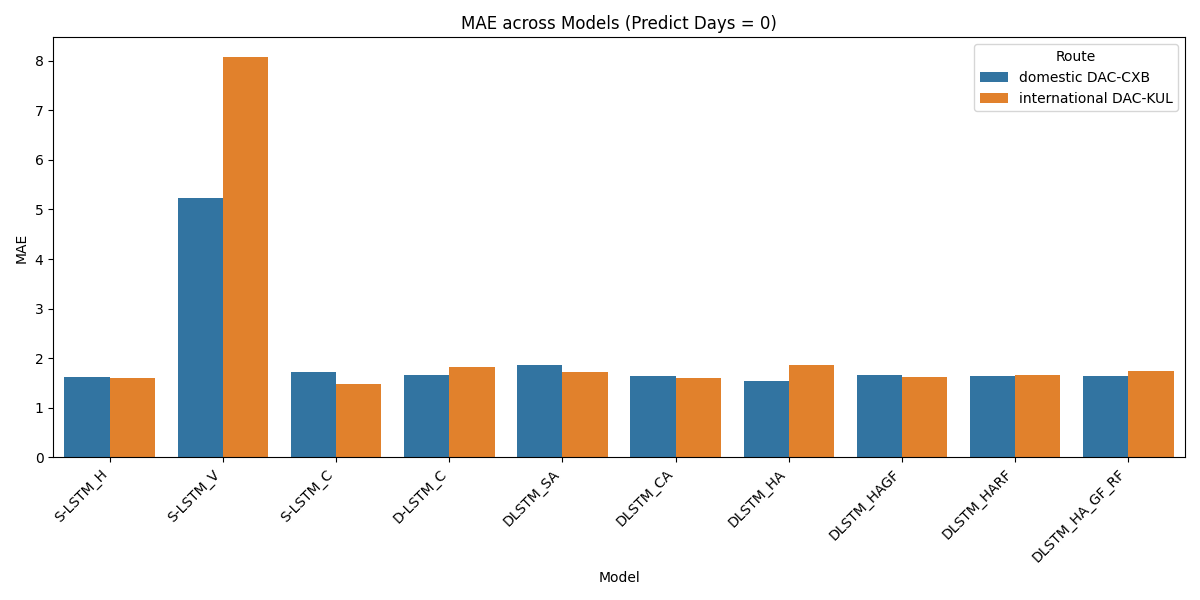}
  \caption{MAE}
\end{subfigure}
\hfill
\begin{subfigure}{0.45\textwidth}
  \centering
  \includegraphics[width=\linewidth]{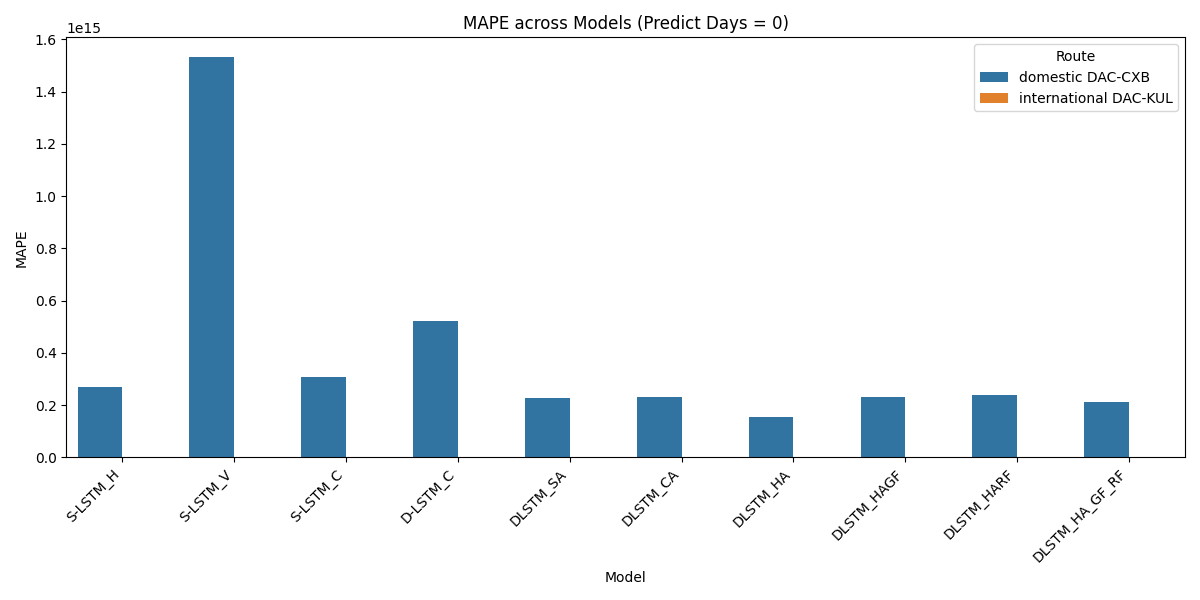}
  \caption{MAPE}
\end{subfigure}

% \vspace{-0.6em}

\begin{subfigure}{0.45\textwidth}
  \centering
  \includegraphics[width=\linewidth]{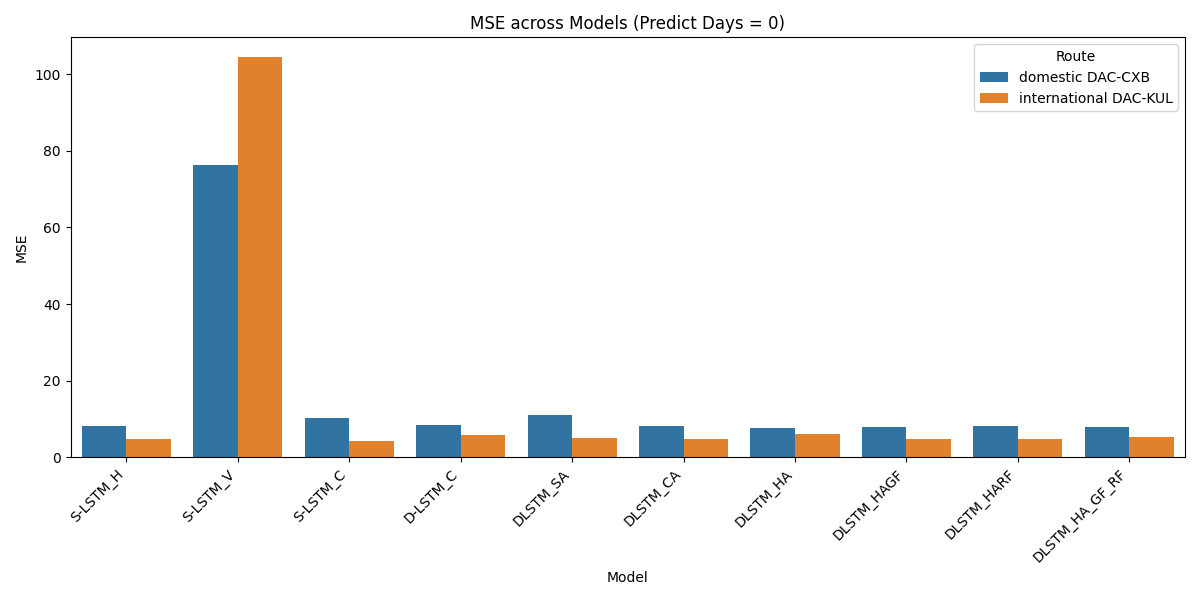}
  \caption{RMSE}
\end{subfigure}
\hfill
\begin{subfigure}{0.45\textwidth}
  \centering
  \includegraphics[width=\linewidth]{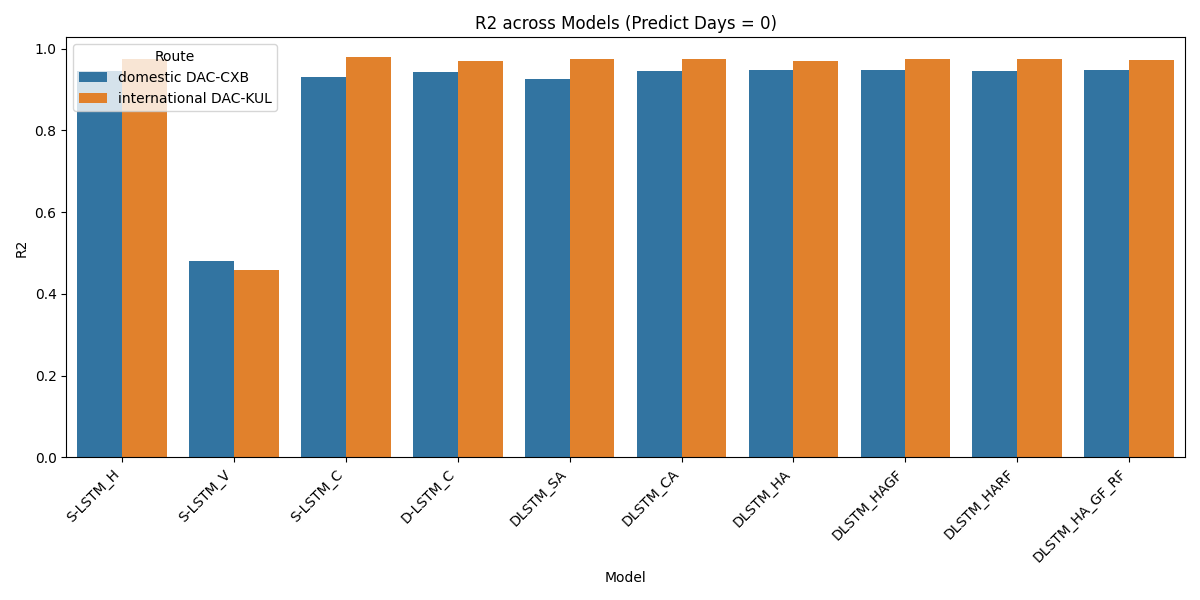}
  \caption{$R^{2}$}
\end{subfigure}

\caption{Metric values for Domestic vs. International flights.}
\label{fig:dom_vs_int}
\end{figure*}

\begin{figure*}[!t]
\centering
\begin{subfigure}{0.45\textwidth}
  \centering
  \includegraphics[width=\linewidth]{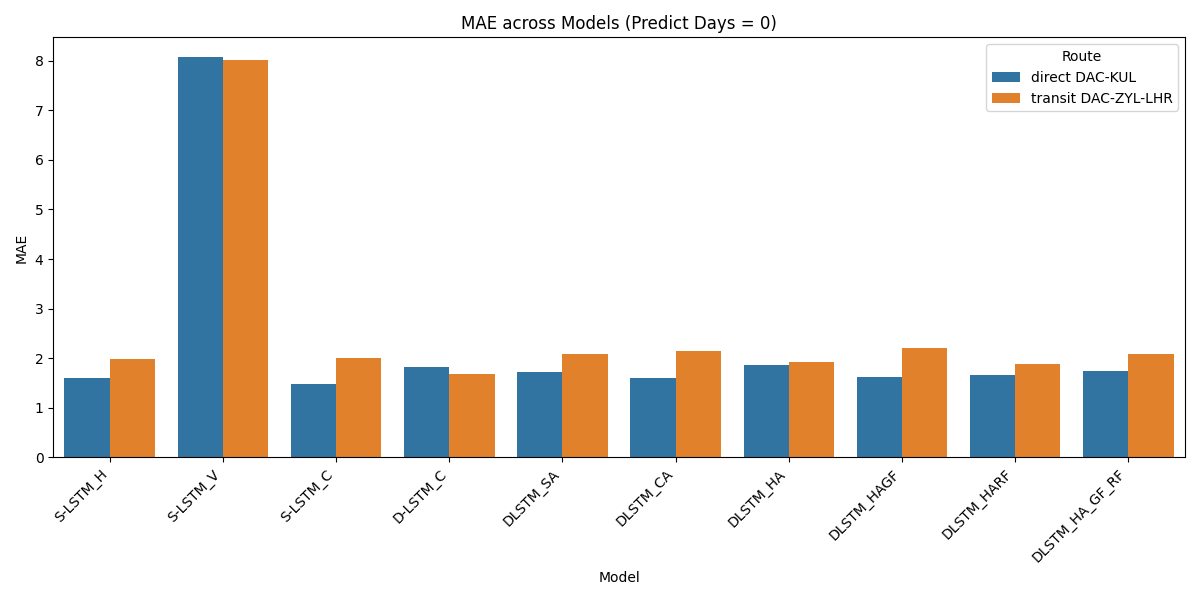}
  \caption{MAE}
\end{subfigure}
\hfill
\begin{subfigure}{0.45\textwidth}
  \centering
  \includegraphics[width=\linewidth]{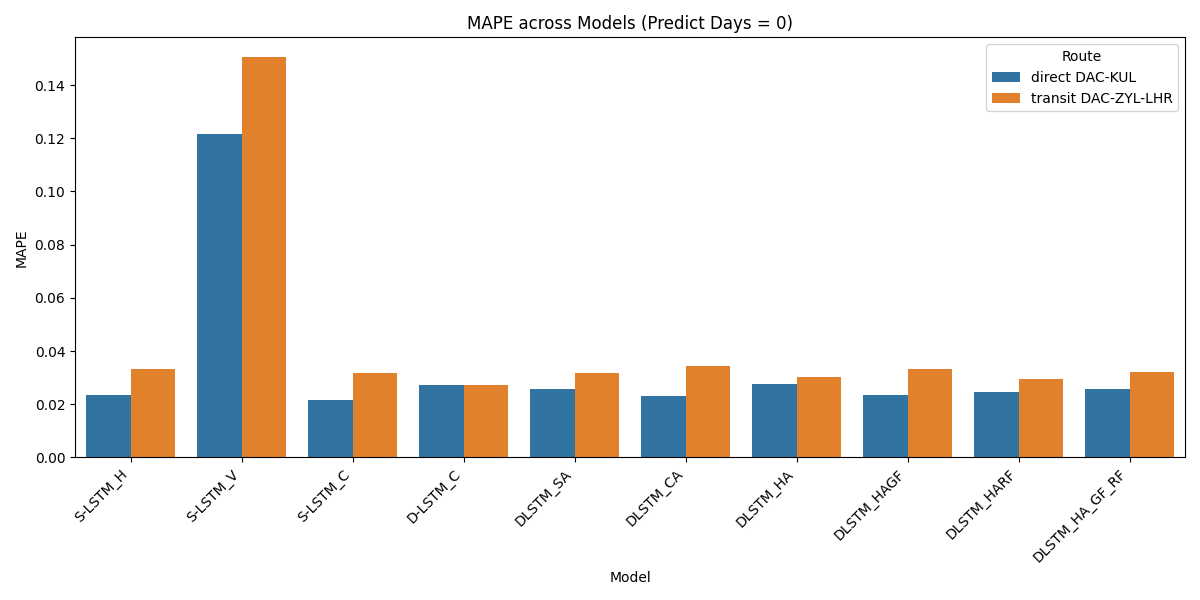}
  \caption{MAPE}
\end{subfigure}

% \vspace{0.8em}

\begin{subfigure}{0.45\textwidth}
  \centering
  \includegraphics[width=\linewidth]{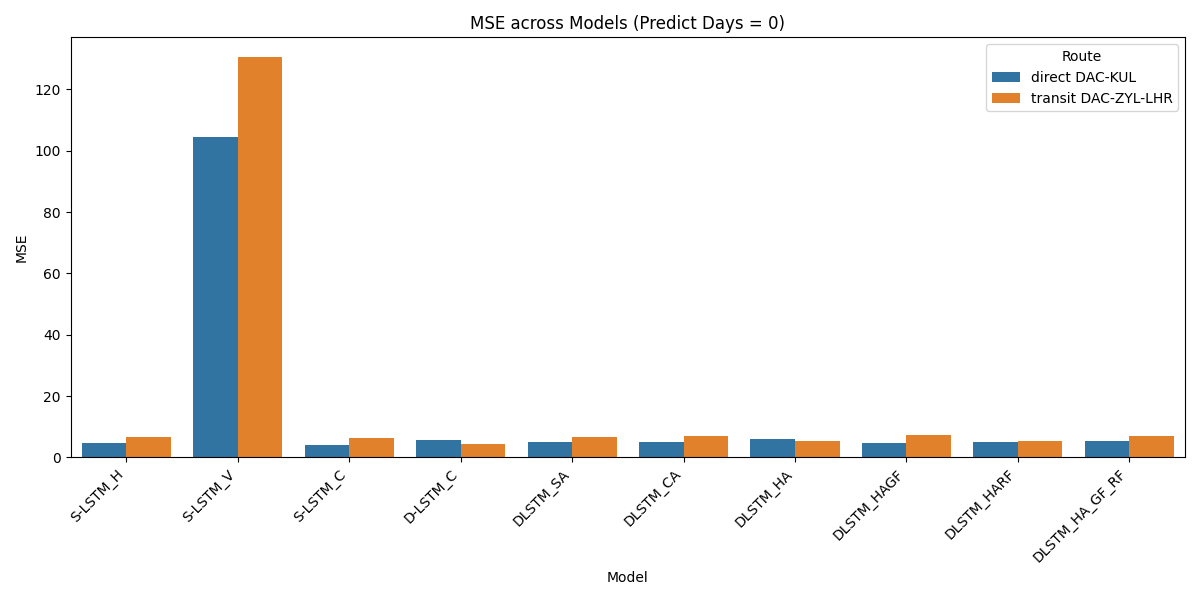}
  \caption{RMSE}
\end{subfigure}
\hfill
\begin{subfigure}{0.45\textwidth}
  \centering
  \includegraphics[width=\linewidth]{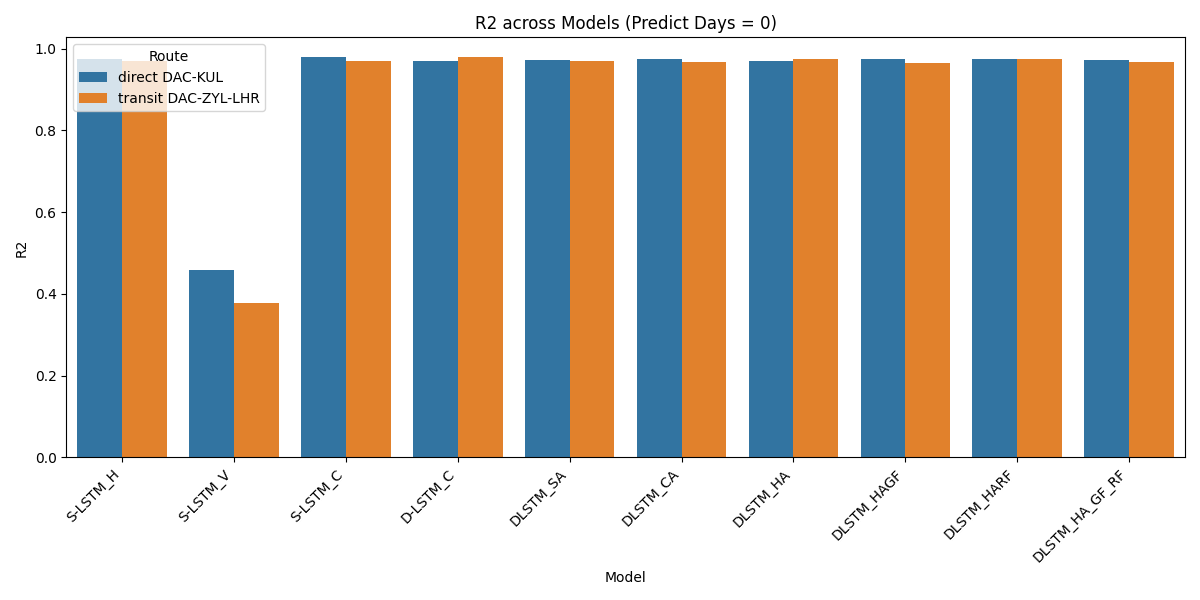}
  \caption{$R^{2}$}
\end{subfigure}

\caption{Metric values for Direct vs. Transit flights.}
\label{fig:direct_vs_transit}
\end{figure*}

\begin{figure*}[!t]
\centering
\begin{subfigure}{0.45\textwidth}
  \centering
  \includegraphics[width=\linewidth]{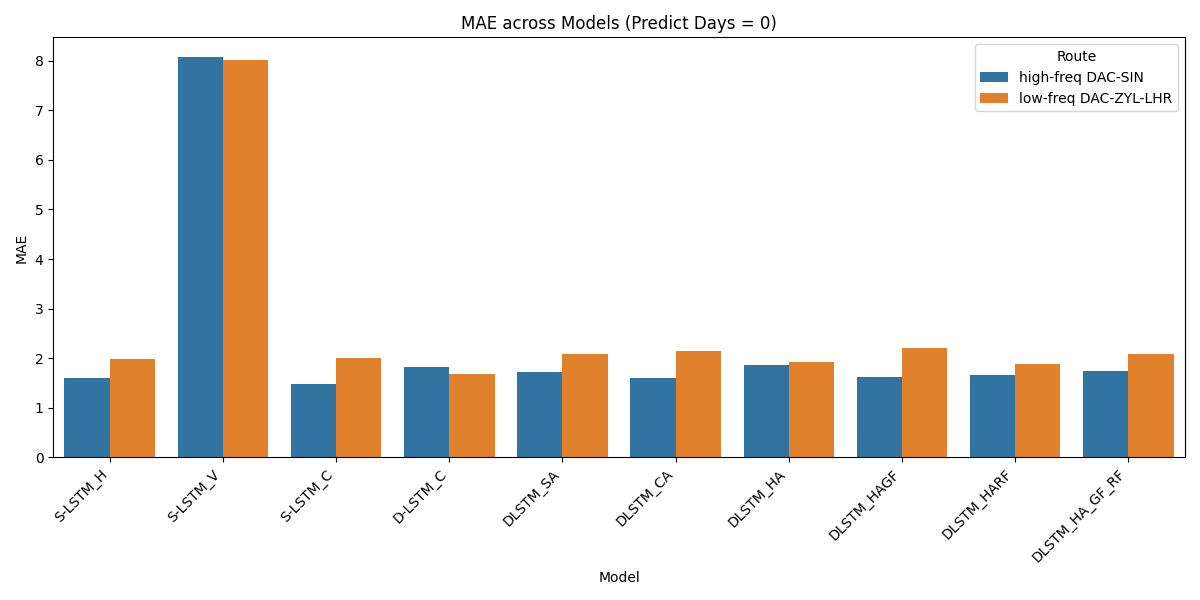}
  \caption{MAE}
\end{subfigure}
\hfill
\begin{subfigure}{0.45\textwidth}
  \centering
  \includegraphics[width=\linewidth]{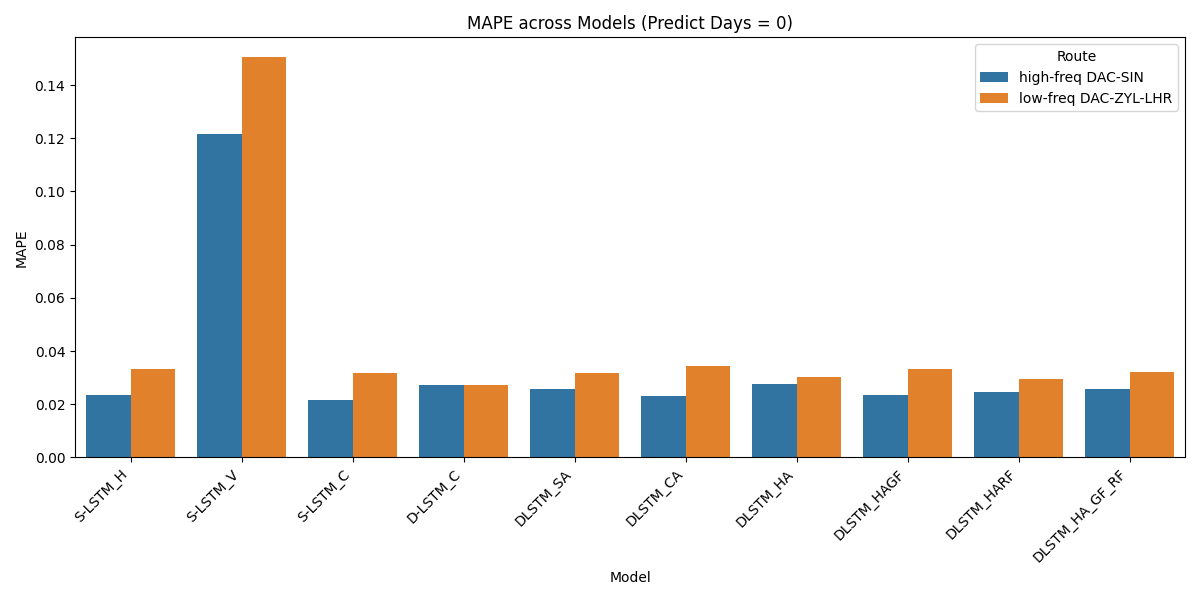}
  \caption{MAPE}
\end{subfigure}

% \vspace{0.8em}

\begin{subfigure}{0.45\textwidth}
  \centering
  \includegraphics[width=\linewidth]{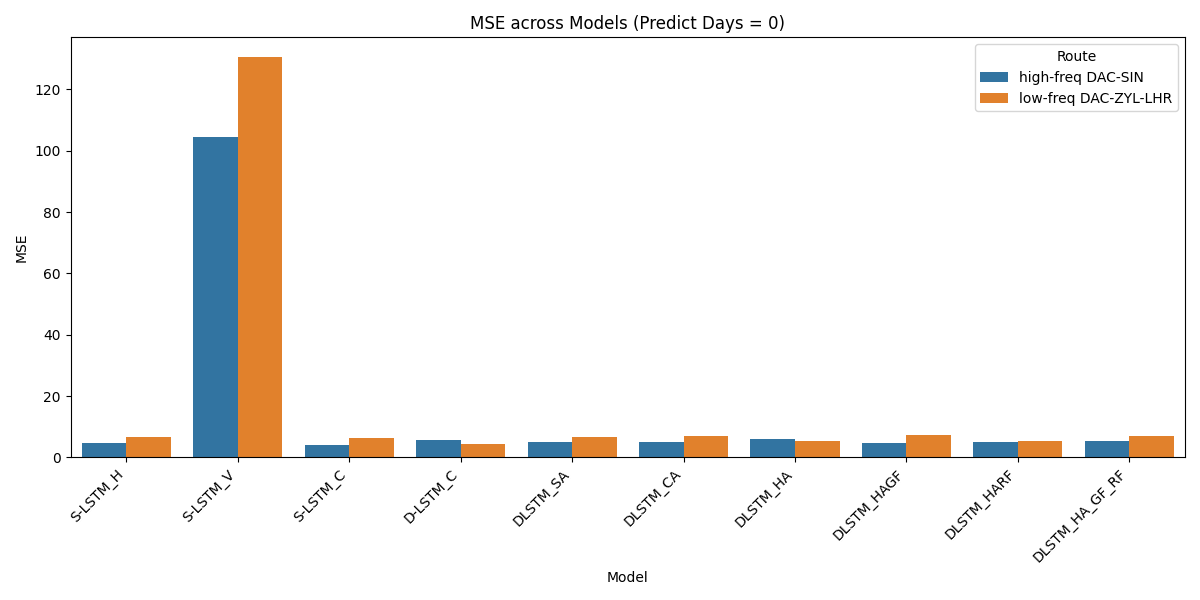}
  \caption{RMSE}
\end{subfigure}
\hfill
\begin{subfigure}{0.45\textwidth}
  \centering
  \includegraphics[width=\linewidth]{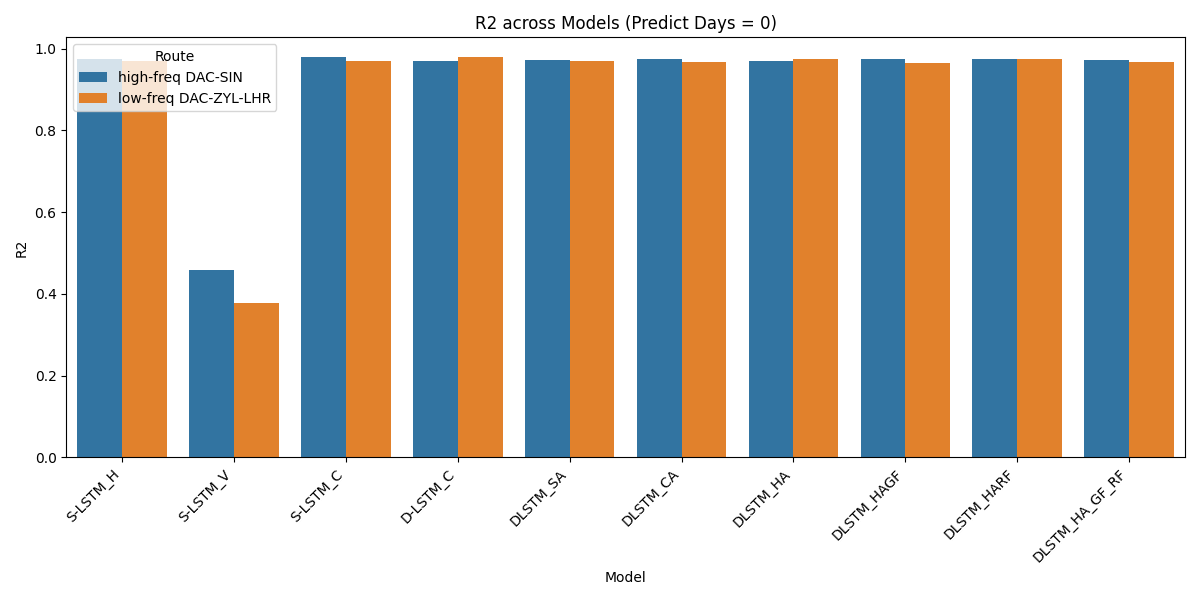}
  \caption{$R^{2}$}
\end{subfigure}

\caption{Metric values for High vs. Low frequency flights.}
\label{fig:freq}
\end{figure*}

\begin{figure*}[!t]
\centering
\begin{subfigure}{0.45\textwidth}
  \centering
  \includegraphics[width=\linewidth]{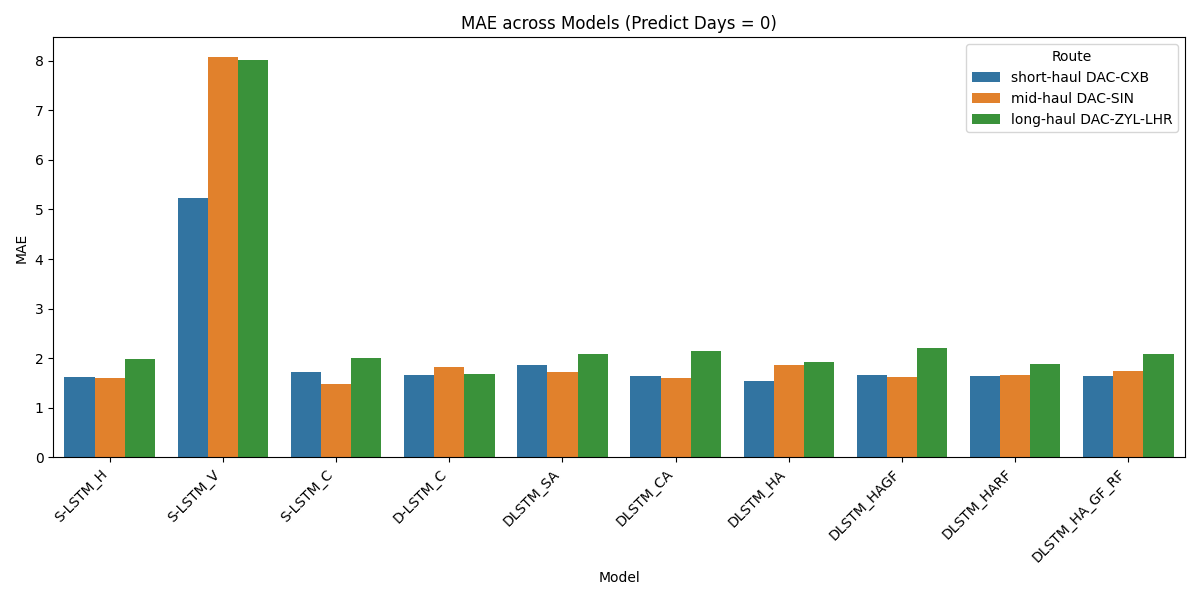}
  \caption{MAE}
\end{subfigure}
\hfill
\begin{subfigure}{0.45\textwidth}
  \centering
  \includegraphics[width=\linewidth]{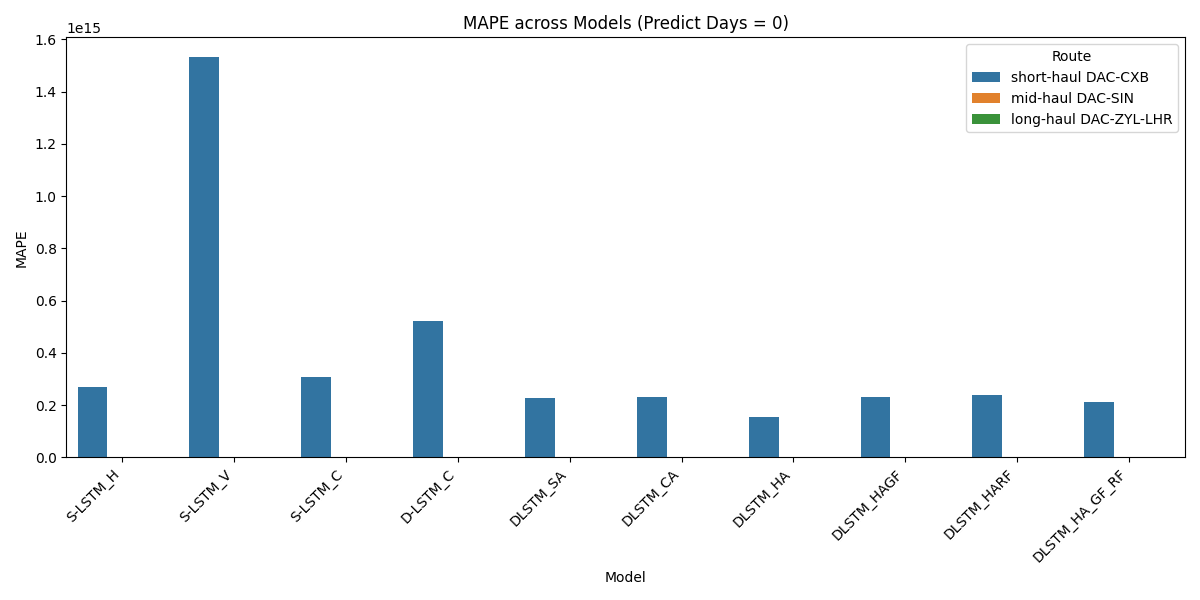}
  \caption{MAPE}
\end{subfigure}

% \vspace{0.8em}

\begin{subfigure}{0.45\textwidth}
  \centering
  \includegraphics[width=\linewidth]{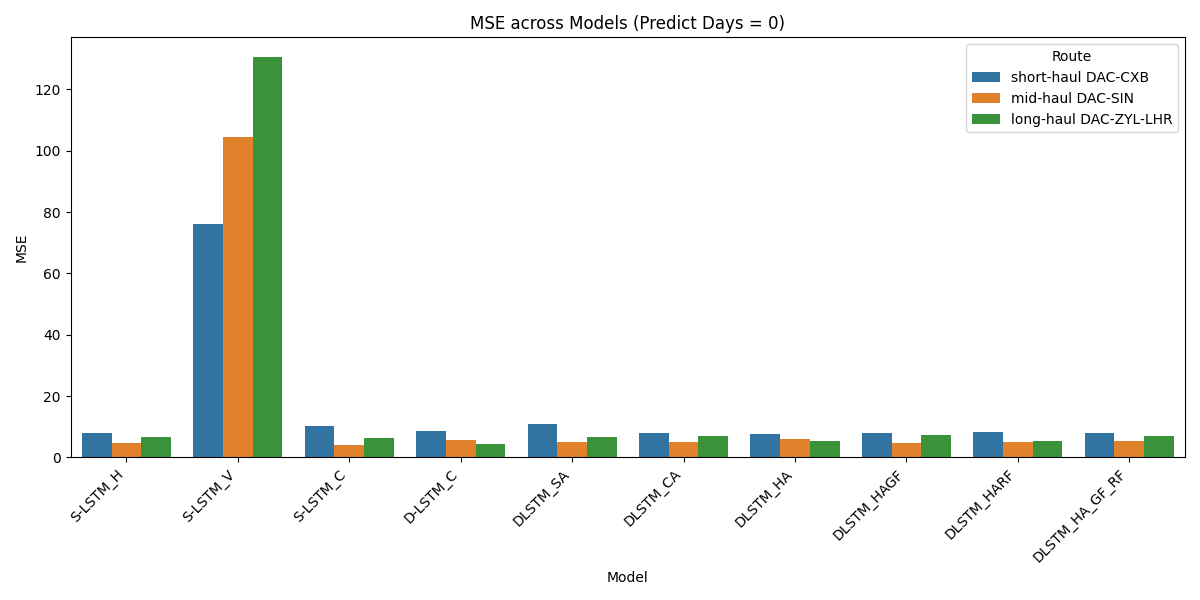}
  \caption{RMSE}
\end{subfigure}
\hfill
\begin{subfigure}{0.45\textwidth}
  \centering
  \includegraphics[width=\linewidth]{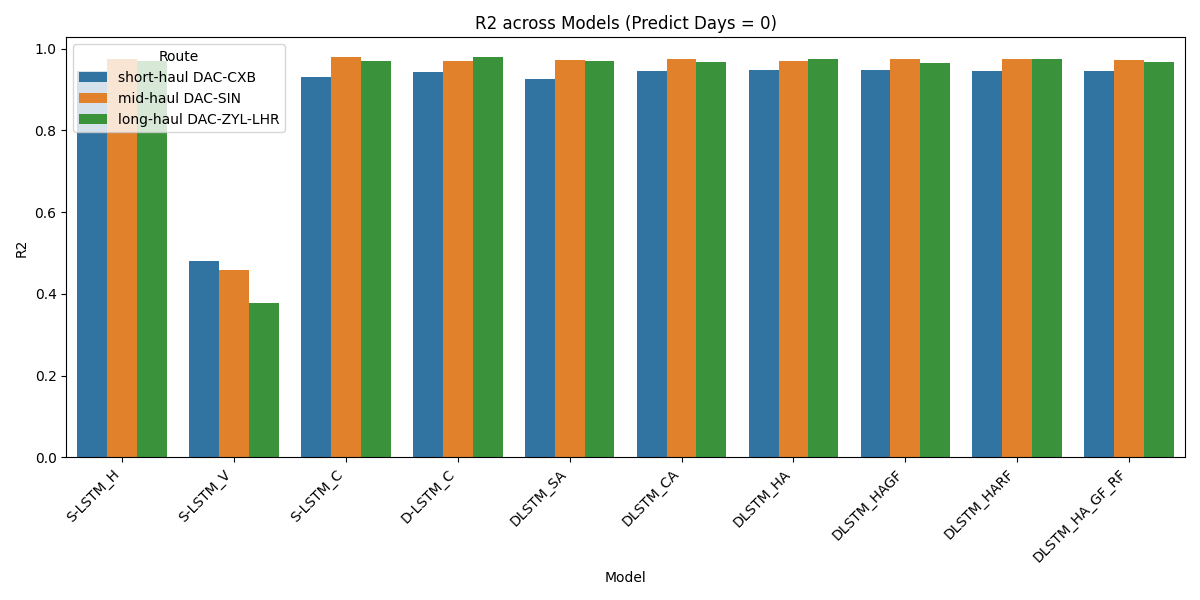}
  \caption{$R^{2}$}
\end{subfigure}

\caption{Metric values for Short vs. Mid vs. Long Haul flights.}
\label{fig:haul}
\end{figure*}

To assess whether the proposed model generalises across all distinct
flight types, DLSTM-HA was evaluated on four route category pairs. The
full per-category metric plots are provided in
Figs.~\ref{fig:dom_vs_int}--\ref{fig:haul}.

\emph{Domestic versus International:} DLSTM-HA maintained consistent
performance across both domestic and international routes.
Vertical-only models underperformed on domestic routes because
domestic routes have shorter booking windows and more last-minute
changes, which weakens the vertical pattern. All models had higher
MAPE on domestic routes, but this was primarily a mathematical issue
caused by near-zero booking counts at the early stages of the booking
window, not a model failure. Cross-attention partially compensates for
the weaker vertical sequence by dynamically reducing its importance
when the horizontal sequence is more informative.

\emph{Direct versus Transit:} Transit flights showed a characteristic
double-peak booking pattern: an early surge from travellers securing
connections in advance, followed by a secondary surge as departure
approaches. Attention-based models, particularly DLSTM-HA, handled
this structure better than concatenation-based baselines.
Cross-attention allows the model to identify and up-weight the most
relevant historical booking positions for transit flights, which
structurally differ from direct flight booking curves.

\emph{High versus Low Frequency:} High-frequency routes demonstrated
more stable and predictable booking patterns, reflected by reduced
error variance across all models. In contrast, low-frequency routes
exhibited greater variability, as idiosyncratic demand fluctuations
are amplified due to fewer opportunities for passenger rebooking or
itinerary adjustment. DLSTM-HA maintained the lowest prediction error
across both frequency segments.

\emph{Short, Mid, and Long Haul:} Long-haul flights showed the most
uniform PLF booking curves due to structured advance-purchase
behaviour, and DLSTM-HA performed consistently well on these routes.
Short-haul flights yielded higher MAPE values like the domestic route
pattern because of high near-zero booking counts early in the booking
window. Despite this, DLSTM-HA maintained the lowest absolute MAE
across all haul categories.

Across all four category pairs, dual-temporal models outperformed
single-stream baselines, confirming that the proposed architecture
generalises across operationally diverse route types.
\begin{table}[!t]
\caption{Comparison With Machine Learning Baselines}
\label{tab:ml}
\centering
\renewcommand{\arraystretch}{1.2}
\begin{tabular}{lccccc}
\toprule
\textbf{Metric} & \textbf{LR} & \textbf{RF} & \textbf{XGBoost} & \textbf{LightGBM} & \makecell{\textbf{DLSTM-HA}\\\textbf{(proposed)}} \\
\midrule
MAE     & 7.25   & 3.32  & 3.61  & 3.29           & \textbf{2.82} \\
MASE    & 1.31   & 0.60  & 0.65  & 0.59           & \textbf{0.56} \\
MAPE    & 0.18   & 0.08  & 0.10  & 0.08           & 0.09          \\
MSE     & 575.83 & 24.67 & 27.73 & \textbf{24.37} & 26.53         \\
RMSE    & 23.99  & 4.96  & 5.26  & \textbf{4.93}  & 5.15          \\
$R^{2}$ & $-$0.21 & 0.94 & 0.94 & 0.94           & \textbf{0.95} \\
\bottomrule
\end{tabular}
\end{table}

\subsection{Comparison with Baseline Models}

Table~\ref{tab:ml} presents the performance of DLSTM-HA alongside
four non-sequential machine learning baselines trained on the same
feature set.

Linear Regression failed substantially, producing negative
$R^{2}\,(=\,{-}0.21)$ and MAE\,$=\,7.25$. This confirms that the
linear assumption is fundamentally inappropriate for PLF prediction
from booking sequences. Random Forest (MAE\,$=\,3.32$,
$R^{2}\,=\,0.94$) and LightGBM (MAE\,$=\,3.29$, $R^{2}\,=\,0.94$)
performed considerably better. Which demonstrates that the engineered
features carry genuine predictive signal even without temporal
modelling. However, both fell short of DLSTM-HA across MAE and MASE.
The performance gap was most evident in MAE (0.47 absolute reduction
compared to LightGBM) and MASE (0.59 reduced to 0.56), underscoring
the importance of modelling sequential structure.

However, LightGBM achieved a slightly lower MSE (24.37) than
DLSTM-HA (26.53). This reflects a specific trade-off: tree-based
models tend to avoid large individual errors on well-observed samples,
while LSTM-based models are trained to minimize MSE globally across
the full booking curve. The LSTM's lower MAE and higher $R^{2}$
indicate superior average-case accuracy and better overall variance
explanation.

\begin{table}[!t]
\caption{Comparison With Prior Airline Demand Prediction Models}
\label{tab:prior}
\centering
\renewcommand{\arraystretch}{1.2}
\begin{tabular}{lcccccccc}
\toprule
\multirow{2}{*}{\textbf{Metric}} &
  \multicolumn{2}{c}{\textbf{Pan \emph{et al.}~\cite{pan2018}}} &
  \multicolumn{2}{c}{\textbf{He \emph{et al.}~\cite{he2023}}} &
  \multicolumn{3}{c}{\textbf{This Paper}} \\
\cmidrule(lr){2-3}\cmidrule(lr){4-5}\cmidrule(lr){6-8}
 & H & V & H & V & H & V & H+V \\
\midrule
MAE     & 3.15  & 12.91  & 3.39  & 12.82  & 2.96  & 11.76  & \textbf{2.82} \\
MASE    & 0.58  & 2.38   & 0.59  & 2.21   & 0.59  & 2.33   & \textbf{0.56} \\
MAPE    & 0.08  & 0.39   & 0.07  & 0.30   & 0.10  & 0.41   & 0.09          \\
MSE     & 28.90 & 250.14 & 28.28 & 262.17 & 26.82 & 231.97 & \textbf{26.53}\\
RMSE    & 5.38  & 15.82  & 5.32  & 16.19  & 5.18  & 15.23  & 5.15          \\
$R^{2}$ & 0.94  & 0.49   & 0.94  & 0.41   & 0.95  & 0.56   & \textbf{0.95} \\
\bottomrule
\end{tabular}
\end{table}

\subsection{Comparison with Prior Studies}

Table~\ref{tab:prior} presents a comparison with
Pan~\emph{et~al.}~\cite{pan2018} and He~\emph{et~al.}~\cite{he2023},
both reimplemented and evaluated on the same dataset under identical
evaluation criteria. DLSTM-HA outperforms both prior models in the horizontal-only configuration (MAE: 2.96 vs.\ 3.15 and 3.39) and matches or improves on their vertical-only configurations. More importantly, the H+V
dual-stream configuration achieved MAE\,$=\,2.82$, a result that
neither prior study could produce because neither integrates both
streams. The $R^{2}$ improvement is modest (0.95 vs.\ 0.94) but
consistent, and the MASE improvement (0.56 vs.\ 0.58--0.59) confirms
that the gains are driven by architectural advancement rather than
dataset characteristics.

\begin{table}[!t]
\caption{Comparison With Prior Dual-LSTM Models}
\label{tab:duallstm}
\centering
\renewcommand{\arraystretch}{1.2}
\begin{tabular}{lcccc}
\toprule
\textbf{Metric} & \textbf{Islam~\cite{islam2021}} & \textbf{Peng~\cite{peng2022}} & \textbf{Gu~\cite{gu2022}} & \makecell{\textbf{DLSTM-HA}\\ \textbf{(proposed)}} \\
\midrule
MAE     & 3.58  & 7.68   & 3.61  & \textbf{2.82} \\
MASE    & 0.65  & 1.39   & 0.65  & \textbf{0.56} \\
MAPE    & 0.09  & 0.20   & 0.10  & 0.09          \\
MSE     & 29.65 & 116.49 & 30.56 & \textbf{26.53}\\
RMSE    & 5.44  & 10.79  & 5.53  & \textbf{5.15} \\
$R^{2}$ & 0.94  & 0.75   & 0.94  & \textbf{0.95} \\
\bottomrule
\end{tabular}
\end{table}

%\subsection{Comparison with Prior Dual-LSTM Models}

Table~\ref{tab:duallstm} presents the comparison with three dual-LSTM
models reimplemented on the same airline booking dataset.

All three prior dual-LSTM models underperformed compared to DLSTM-HA.
The model by Peng~\emph{et~al.}~\cite{peng2022} yielded the weakest
result (MAE\,$=\,7.68$, $R^{2}\,=\,0.75$), reflecting the
architectural mismatch between their sensor-based time series with
slowly evolving features and the relatively short, volatile booking
sequences used here. Islam~\emph{et~al.}~\cite{islam2021} and
Gu~\emph{et~al.}~\cite{gu2022} achieved closer performance
(MAE\,$=\,3.58$ and 3.61 respectively), but both remained well above
DLSTM-HA's MAE of 2.82. This demonstrates that dual-stream
architectures developed for other contexts do not effectively transfer
to airline booking forecasting without modification. The proposed
hybrid attention mechanism, specifically designed to model the
complementary intra-flight and inter-flight sequences, confers a
notable architectural advantage.

% ============================================================
%  VI. CONCLUSION
% ============================================================
\section{Conclusion}
\label{sec:conclusion}

%\subsection{Summary of Contributions}

This paper addressed three interconnected limitations in the existing
airline demand forecasting literature: the unidimensional view of
booking sequences, the operational fragility of absolute passenger
count as a forecasting target, and the absence of a structured feature
selection framework tailored to dual-temporal reservation data. To address these gaps, a dual-stream LSTM framework was proposed that processes horizontal and vertical booking sequences through separate parallel branches. The horizontal stream captures how bookings accumulate for a given flight as its departure approaches, while the vertical stream captures how similar flights on the same route
historically behaved at the same days-before-departure position. A
novel offset-based algorithm ensures that each historical reference
occupies a structurally equivalent position on the booking curve. The
framework targets PLF rather than absolute passenger counts, and a
seven-stage feature selection pipeline reduces 39 engineered features
to 17 optimally informative inputs. The hybrid attention model
(DLSTM-HA) achieved the best overall performance, with
MAE\,$=\,2.8167$ and $R^{2}\,=\,0.9495$. The model was validated
across four operationally distinct flight category pairs. The
proposed methodology has been officially adopted by Biman Bangladesh Airlines (BBA).

%\subsection{Key Findings}

First, the vertical sequence alone is a weak predictor of PLF in
isolation ($R^{2}\,=\,0.56$). However, when properly integrated
through cross-attention, the vertical signal meaningfully improved
performance beyond what the horizontal stream alone could achieve.
This was evident on domestic and low-frequency flight categories where
booking patterns are volatile. Second, the choice of fusion mechanism matters. Simply concatenating horizontal and vertical features (SLSTM-C) produced marginal
improvement over the horizontal-only baseline. Separate dual-stream
processing without attention (DLSTM) did not yield a clear advantage.
It was only when cross-attention was introduced a consistent and
meaningful performance gain over all single-stream baselines was
achieved.  Third, the dynamic horizon analysis revealed that the relationship
between prediction horizon and forecast accuracy is not monotonic.
Performance fluctuated without a clear trend during the mid-range
booking window (D-21 to approximately D-8), then degraded consistently
in the final week before departure due to last-minute cancellations,
upgrades, and no-shows. Dual-temporal models showed the most stable
behaviour across all horizons. Fourth, shifting the forecasting target from absolute passenger counts to PLF produced a practically important generalisation benefit. The
same trained model produced valid and interpretable predictions across
all flight categories, including routes where equipment changes altered
total capacity.

%\subsection{Limitations and Future Directions}

%This study has a few limitations that can be addressed in future work. The dataset is drawn from a single carrier operating primarily within the Bangladesh context. Extending the framework to multi-airline or multi-country datasets would strengthen the generalisability claim considerably. The current model predicts aggregate PLF across all cabin classes; Extending the proposed framework to produce separate class-wise PLF forecasts would make it directly usable in cabin-level revenue management systems and represents a natural next step. Finally, the study does not incorporate competitive market signals such as competitor load factors, fare dynamics, or schedule changes on the same route. Incorporating such information as additional vertical features or as a third input stream is a meaningful direction for extending the current architecture.

This study has a few limitations that can be addressed in future work. The dataset is drawn from a single carrier operating primarily within the Bangladesh context. Extending the framework to multi-airline or multi-country datasets would strengthen the generalisability of our claim. The current model predicts aggregate PLF across all cabin classes; Extending the proposed framework to produce separate class-wise PLF forecasts would make it directly usable in cabin-level revenue management systems and represents a natural next step. 

% ============================================================
%  ACKNOWLEDGMENT
% ============================================================
\section*{Acknowledgment}
We are grateful to IICT BUET for allowing us to use its all kinds of facilities. The authors would like to express gratitude to Biman Bangladesh
Airlines for providing access to real-world operational data.

\section*{Declaration of Generative AI Use}
During the preparation of this manuscript, the author used AI-assisted language tools in order to assist with language editing, grammar correction, and improving the readability of the manuscript. After using these tools/services, the author reviewed and edited the content as needed and take full responsibility for the content of the published article.

% ============================================================
%  REFERENCES
% ============================================================
\bibliographystyle{IEEEtran}
\bibliography{bibliography}

@misc{iata2025,
  title        = {Airline Profitability to Strengthen Slightly in 2025 Despite Headwinds},
  howpublished = {\url{https://www.iata.org/en/pressroom/2025-releases/2025-06-02-01/}},
  note         = {Accessed: Jul.~18, 2025},
  year         = {2025},
}

@article{nieto2018,
  author  = {Nieto, M. R. and Carmona-Ben{\'{i}}tez, R. B.},
  title   = {{ARIMA + GARCH + Bootstrap} forecasting method applied to the airline industry},
  journal = {Journal of Air Transport Management},
  volume  = {71},
  pages   = {1--8},
  year    = {2018},
  doi     = {10.1016/j.jairtraman.2018.05.007},
}

@article{suryan2017,
  author  = {Suryan, V.},
  title   = {Econometric forecasting models for air traffic passenger of {Indonesia}},
  journal = {Journal of Civil Engineering Forum},
  volume  = {3},
  number  = {1},
  year    = {2017},
  doi     = {10.22146/jcef.26594},
}

@inproceedings{pan2018,
  author    = {Pan, B. and Yuan, D. and Sun, W. and Liang, C. and Li, D.},
  title     = {A novel {LSTM}-based daily airline demand forecasting method using vertical and horizontal time series},
  booktitle = {Trends and Applications in Knowledge Discovery and Data Mining (PAKDD Workshops)},
  pages     = {168--173},
  publisher = {Springer},
  address   = {Cham},
  year      = {2018},
  doi       = {10.1007/978-3-030-04503-6_17},
}

@article{he2023,
  author  = {He, H. and Chen, L. and Wang, S.},
  title   = {Flight short-term booking demand forecasting based on a long short-term memory network},
  journal = {Computers \& Industrial Engineering},
  volume  = {186},
  pages   = {109707},
  year    = {2023},
  doi     = {10.1016/j.cie.2023.109707},
}

@article{carmona2020,
  author  = {Carmona-Ben{\'{i}}tez, R. B. and Nieto, M. R.},
  title   = {{SARIMA} damp trend grey forecasting model for airline industry},
  journal = {Journal of Air Transport Management},
  volume  = {82},
  pages   = {101736},
  year    = {2020},
  doi     = {10.1016/j.jairtraman.2019.101736},
}

@article{wu2021,
  author  = {Wu, X. and Xiang, Y. and Mao, G. and Du, M. and Yang, X. and Zhou, X.},
  title   = {Forecasting air passenger traffic flow based on the two-phase learning model},
  journal = {Journal of Supercomputing},
  volume  = {77},
  number  = {5},
  pages   = {4221--4243},
  year    = {2021},
  doi     = {10.1007/s11227-020-03428-2},
}

@article{grimme2020,
  author  = {Grimme, W. and Bingemer, S. and Maertens, S.},
  title   = {An analysis of the prospects of ultra-long-haul airline operations using passenger demand data},
  journal = {Transportation Research Procedia},
  volume  = {51},
  pages   = {208--216},
  year    = {2020},
  doi     = {10.1016/j.trpro.2020.11.023},
}

@article{dey2022,
  author  = {Dey Tirtha, S. and Bhowmik, T. and Eluru, N.},
  title   = {An airport level framework for examining the impact of {COVID-19} on airline demand},
  journal = {Transportation Research Part A: Policy and Practice},
  volume  = {159},
  pages   = {169--181},
  year    = {2022},
  doi     = {10.1016/j.tra.2022.03.014},
}

@article{bastola2017,
  author  = {Bastola, D. P.},
  title   = {Air Passenger Demand Model ({APDM}): Econometric model for forecasting demand in passenger air transports in {Nepal}},
  volume  = {1},
  number  = {4},
  year    = {2017},
}

@article{haensel2011,
  author  = {Haensel, A. and Koole, G. and Erdman, J.},
  title   = {Estimating unconstrained customer choice set demand: {A} case study on airline reservation data},
  journal = {Journal of Choice Modelling},
  volume  = {4},
  number  = {3},
  pages   = {75--87},
  year    = {2011},
  doi     = {10.1016/S1755-5345(13)70043-5},
}

@unpublished{vanostaijen,
  author = {van Ostaijen, T. and Santos, B. F. and Mitici, M.},
  title  = {Dynamic airline booking forecasting},
  note   = {Unpublished manuscript},
}

@mastersthesis{marques2016,
  author = {Marques, T. L. E.},
  title  = {Dynamic airline booking demand forecasting},
  year   = {2016},
}

@article{firat2021,
  author  = {Firat, M. and Yiltas-Kaplan, D. and Samli, R.},
  title   = {Forecasting air travel demand for selected destinations using machine learning methods},
  journal = {Journal of Universal Computer Science},
  volume  = {27},
  number  = {6},
  year    = {2021},
  doi     = {10.3897/jucs.68185},
}

@article{chen2020,
  author  = {Chen, J.-H. and Wei, H.-H. and Chen, C.-L. and Wei, H.-Y. and Chen, Y.-P. and Ye, Z.},
  title   = {A practical approach to determining critical macroeconomic factors in air-traffic volume based on {K}-means clustering and decision-tree classification},
  journal = {Journal of Air Transport Management},
  volume  = {82},
  pages   = {101743},
  year    = {2020},
  doi     = {10.1016/j.jairtraman.2019.101743},
}

@article{hopman2021,
  author  = {Hopman, D. and Koole, G. and Mei, R. V. D.},
  title   = {A machine learning approach to itinerary-level booking prediction in competitive airline markets},
  journal = {International Journal of Revenue Management},
  volume  = {12},
  number  = {3--4},
  pages   = {153--191},
  year    = {2021},
  doi     = {10.1504/IJRM.2021.120347},
}

@article{ghandeharioun2023,
  author  = {Ghandeharioun, Z. and Zendehdel Nobari, P. and Wu, W.},
  title   = {Exploring deep learning approaches for short-term passenger demand prediction},
  journal = {Data Science for Transportation},
  volume  = {5},
  number  = {3},
  pages   = {19},
  year    = {2023},
  doi     = {10.1007/s42421-023-00075-w},
}

@incollection{rumelhart1987,
  author    = {Rumelhart, D. E. and McClelland, J. L.},
  title     = {Learning internal representations by error propagation},
  booktitle = {Parallel Distributed Processing: Explorations in the Microstructure of Cognition},
  publisher = {MIT Press},
  year      = {1987},
  pages     = {318--362},
}

@article{hochreiter1997,
  author  = {Hochreiter, S. and Schmidhuber, J.},
  title   = {Long short-term memory},
  journal = {Neural Computation},
  volume  = {9},
  number  = {8},
  pages   = {1735--1780},
  year    = {1997},
  doi     = {10.1162/neco.1997.9.8.1735},
}

@article{do2020,
  author  = {Do, Q. H. and Lo, S.-K. and Chen, J.-F. and Le, C.-L. and Anh, L. H.},
  title   = {Forecasting air passenger demand: a comparison of {LSTM} and {SARIMA}},
  journal = {Journal of Computer Science},
  volume  = {16},
  number  = {7},
  pages   = {1063--1084},
  year    = {2020},
}

@article{kanavos2021,
  author  = {Kanavos, A. and Kounelis, F. and Iliadis, L. and Makris, C.},
  title   = {Deep learning models for forecasting aviation demand time series},
  journal = {Neural Computing and Applications},
  volume  = {33},
  number  = {23},
  pages   = {16329--16343},
  year    = {2021},
  doi     = {10.1007/s00521-021-06232-y},
}

@article{iacus2020,
  author  = {Iacus, S. M. and Natale, F. and Santamaria, C. and Spyratos, S. and Vespe, M.},
  title   = {Estimating and projecting air passenger traffic during the {COVID-19} coronavirus outbreak and its socio-economic impact},
  journal = {Safety Science},
  volume  = {129},
  pages   = {104791},
  year    = {2020},
  doi     = {10.1016/j.ssci.2020.104791},
}

@inproceedings{islam2021,
  author    = {Islam, Z. and Rukonuzzaman, M. and Ahmed, R. and Kabir, Md. H. and Farazi, M.},
  title     = {Efficient two-stream network for violence detection using separable convolutional {LSTM}},
  booktitle = {Proc.\ International Joint Conference on Neural Networks (IJCNN)},
  pages     = {1--8},
  year      = {2021},
  doi       = {10.1109/IJCNN52387.2021.9534280},
}

@article{gu2022,
  author  = {Gu, Y. H. and Jin, D. and Yin, H. and Zheng, R. and Piao, X. and Yoo, S. J.},
  title   = {Forecasting agricultural commodity prices using dual input attention {LSTM}},
  journal = {Agriculture},
  volume  = {12},
  number  = {2},
  year    = {2022},
  doi     = {10.3390/agriculture12020256},
}

@article{peng2022,
  author  = {Peng, C. and Wu, J. and Wang, Q. and Gui, W. and Tang, Z.},
  title   = {Remaining useful life prediction using dual-channel {LSTM} with time feature and its difference},
  journal = {Entropy},
  volume  = {24},
  number  = {12},
  year    = {2022},
  doi     = {10.3390/e24121818},
}

@article{simsek2024,
  author  = {{\c{S}}im{\c{s}}ek, K. and Tu{\u{g}}rul, N. {\"O}. {\"O}. and Kara{\c{c}}uha, K. and Tabatadze, V. and Kara{\c{c}}uha, E.},
  title   = {Modeling and predicting passenger load factor in air transportation: {A} deep assessment methodology with fractional calculus approach utilizing reservation data},
  journal = {Fractal and Fractional},
  volume  = {8},
  number  = {4},
  pages   = {214},
  year    = {2024},
  doi     = {10.3390/fractalfract8040214},
}

\end{document}